# Variance Control via *Weight Rescaling* in LLM Pre-training

**Louis Owen**[*], **Abhay Kumar**[*], **Nilabhra Roy Chowdhury, Fabian Güra**
BluOrion
[*]First Authors
{louis.owen, abhay.kumar, nilabhra.chowdhury, fabian.guera}@bluorion.com

March 21, 2025

## Abstract

The outcome of Large Language Model (LLM) pre-training strongly depends on weight initialization and variance control strategies. Although the importance of initial variance control has been well documented in neural networks in general, the literature on initialization and management of its growth during LLM pre-training, specifically, is somewhat sparse. In this paper, we introduce the **Layer Index Rescaling (LIR)** weight initialization scheme, and the **Target Variance Rescaling (TVR)** variance control strategy. Experiments on a 1B parameter LLaMA model demonstrate that better variance management using these techniques yields substantial improvements in downstream task performance (up to 4.6% on common pre-training benchmarks) and reduces extreme activation values, thus mitigating challenges associated with quantization and low-precision training. Our code is available at: https://github.com/bluorion-com/weight_rescaling.

## 1 Introduction

Statistical properties of parameter initialization are the foundation of stable neural network training. In seminal works, researchers have established sophisticated initialization schemes to determine appropriate mean and standard deviation values for weight distributions to ensure stable gradient flow, and hence enable effective training [1, 2]. However, a critical, yet often overlooked phenomenon arises during the training process: weights frequently diverge from their carefully calibrated initial scale, potentially undermining the benefits that proper initialization aims to provide. As training progresses, neural networks gradually experience what recent studies have termed *plasticity loss*–the decreasing capacity to learn new information [3]. A primary cause of this phenomenon is unbounded weight growth, which not only contributes to plasticity loss, but also harms generalization capabilities and disrupts optimization dynamics [4].

Proposed mitigations include *weight decay*, *layer normalization*[5], as well as various weight regularization techniques [6, 7]. In this context, the importance of variance control has been well documented in neural networks generally [8, 9, 10], though for Transformer-based Large Language Models (LLMs), specifically, initial and continuous variance management during pre-training seems to have received limited attention so far. This is particularly problematic given the resource-intensive nature of LLM training, where optimizing every aspect of the process can yield substantial efficiency gains and performance improvements.

The widespread use of normalization techniques such as LayerNorm [11] or RMSNorm [12], or regularization techniques such as weight decay, may have inadvertently caused practitioners to overlook the importance of tailored initialization and rescaling strategies. These normalization techniques mitigate some issues associated with improper scaling by stabilizing activations and gradients, creating a false sense of robustness. However, reliance on normalization alone does not address the underlying problem of weight divergence.

In this paper, we therefore introduce the **Layer Index Rescaling (LIR)** weight initialization scheme (see 2.3.6; inspired by [13]), as well as the **Target Variance Rescaling (TVR)** variance control strategy (see 2.4.3; inspired by [10]). In conjunction with the standard normalization techniques mentioned above, they prevent the weights from deviating too much from their initial distribution throughout the training process. Through systematic tuning and ablation experiments on a 1-billion-parameter LLaMA model, we demonstrate that proper management of weight variance–both at initialization and during training–can yield substantial improvements in downstream task performance while reducing extreme activation values that pose challenges for quantization [14] and low-precision training [15, 16, 17].

## 2 Variance Control Methods

### 2.1 Weight Regularization

#### 2.1.1 L2 regularization

L2 regularization is one of the most widely adopted techniques to constrain weight growth and improve model performance. This approach introduces an additional loss term that penalizes the L2 norm of the weights, discouraging the model from assigning excessively large values to its parameters during training. By imposing this constraint, L2 regularization aims to enhance generalization and robustness while mitigating the risk of overfitting to specific data points or patterns. In the following, $\theta$ denotes the parameters of the neural network, $\mathcal{L}_{\text{data}}(\theta)$ denotes the loss function used to optimize the network with respect to the given training data, $R(\theta)$ is the regularization term, $\lambda$ is a hyperparameter that controls the aggressiveness of the weight decay, and the final loss function $\mathcal{L}(\theta)$ is defined as:

$$\mathcal{L}(\theta) = \mathcal{L}_{\text{data}}(\theta) + \lambda * R(\theta), \tag{1}$$

where $R(\theta) = \|\theta\|_2$.

#### 2.1.2 Weight decay

The terms weight decay and L2 regularization are often used interchangeably in literature, yet they differ subtly in their application, particularly when combined with modern optimization algorithms like Adam or SGD with momentum. L2 regularization directly adds a penalty term to the loss function, while weight decay explicitly modifies the weight update rule during gradient descent. This distinction becomes significant in adaptive optimization methods, where weight decay can lead to different training dynamics compared to traditional L2 regularization. Despite these differences, both techniques share the common goal of controlling the weight magnitudes to improve generalization and stability. However, as models grow larger and more complex, the limitations of these methods have become increasingly apparent. For example, [18] and [19] demonstrated that weight decay may not always effectively control weight growth or maintain desirable properties in certain scenarios, particularly in large-scale models.

#### 2.1.3 Spectral Norm and Rank-Based Regularization

[20] proposed regularizing the spectral norm of the weight matrix. Their experiments demonstrated improved generalization performance across various tasks, showcasing the potential of spectral norm regularization as a complementary approach to traditional weight decay. Similarly, [21] introduced a method to maintain the effective rank of learned features by regularizing the weights. This technique ensures that the model retains diverse and informative representations, preventing the collapse of feature spaces into lower-dimensional subspaces.

#### 2.1.4 Distance-based Regularization

In addition to directly regularizing weight magnitudes, recent studies have explored approaches that leverage the properties of initialized weights to guide training dynamics. [22] proposed imposing a penalty on the L2 distance between the current weights and their initial values. This method encourages the model to stay close to its initialization, thereby preserving the beneficial properties of the initial weight distribution. Similarly, [6] introduced a novel regularization technique based on the empirical Wasserstein distance, which measures the deviation of the weight distribution from its initial state. By minimizing this distance, the approach ensures that the weights remain statistically aligned with their initialization. However, both methods introduce additional computational overhead due to the need for extra gradient computations, making them less practical for resource-intensive models such as LLMs.

## 2.2 Layer Normalization

### 2.2.1 LayerNorm

LayerNorm [11] is a widely adopted normalization technique designed to stabilize the training of deep neural networks by mitigating issues related to internal covariate shift. Unlike Batch Normalization [23], which normalizes activations across a mini-batch, LayerNorm operates on a per-sample basis, making it particularly suitable for tasks such as natural language processing (NLP).

The core idea behind LayerNorm is to normalize the inputs to each layer such that their mean and variance remain consistent across training. Given a hidden representation $\mathbf{h} \in \mathbb{R}^d$ at a particular layer, LayerNorm computes the normalized activations as follows:

$$\mu = \frac{1}{d} \sum_{i=1}^{d} h_i, \quad \sigma^2 = \frac{1}{d} \sum_{i=1}^{d} (h_i - \mu)^2, \tag{2}$$

$$\hat{\mathbf{h}} = \frac{\mathbf{h} - \mu}{\sqrt{\sigma^2 + \epsilon}}, \tag{3}$$

where $\mu$ and $\sigma^2$ represent the mean and variance of the input vector $\mathbf{h}$, respectively, and $\epsilon$ is a small constant added for numerical stability. The normalized activations $\hat{\mathbf{h}}$ are then scaled and shifted using learnable parameters $\boldsymbol{\gamma}, \boldsymbol{\beta} \in \mathbb{R}^d$:

$$\text{LN}(\mathbf{h}) = \boldsymbol{\gamma}\hat{\mathbf{h}} + \boldsymbol{\beta}, \tag{4}$$

where $\boldsymbol{\gamma}$ and $\boldsymbol{\beta}$ allow the model to adaptively re-scale and shift the normalized values.

LayerNorm has proven especially effective in stabilizing the training of large-scale models using Transformer-based architectures [24]. By normalizing across the feature dimension, it reduces sensitivity to weight initialization and helps maintain stable gradient flow during backpropagation. However, while LayerNorm addresses some challenges associated with improper scaling, it does not explicitly constrain the growth of weights themselves.

### 2.2.2 RMSNorm

RMSNorm [12] is a computationally efficient normalization technique that has gained popularity in large-scale neural networks, particularly in Transformer-based architectures. Unlike Layer Normalization (LayerNorm), which normalizes activations using both mean and variance, RMSNorm simplifies the process by omitting the mean subtraction step. This makes it faster to compute while still providing effective stabilization during training.

The key idea behind RMSNorm is to normalize the inputs to each layer based solely on the root mean square (RMS) of the activations. Given a hidden representation $\mathbf{h} \in \mathbb{R}^d$ at a particular layer, RMSNorm computes the normalized activations as follows:

$$\text{RMS}(\mathbf{h}) = \sqrt{\frac{1}{d} \sum_{i=1}^{d} h_i^2 + \epsilon}, \tag{5}$$

$$\hat{\mathbf{h}} = \frac{\mathbf{h}}{\text{RMS}(\mathbf{h})}, \tag{6}$$

where $\epsilon$ is a small constant added for numerical stability. The normalized activations $\hat{\mathbf{h}}$ are then adjusted as:

$$\text{RMSNorm}(\mathbf{h}) = \boldsymbol{\gamma}\hat{\mathbf{h}}, \tag{7}$$

where a learnable parameter $\boldsymbol{\gamma} \in \mathbb{R}^d$ allows the model to adaptively re-scale the normalized values.

Empirical studies have shown that RMSNorm achieves comparable or even superior performance to LayerNorm in many scenarios, especially in Transformer-based architectures. However, similar to LayerNorm, RMSNorm focuses on stabilizing activations rather than explicitly constraining weight growth. This limitation underscores the need for complementary techniques, such as weight rescaling, to manage weight variance throughout training and prevent unbounded growth.

### 2.3 Weight Initialization

#### 2.3.1 Xavier Initialization

Xavier Initialization [1], also known as Glorot Initialization, is a widely used weight initialization technique designed to address the challenges of vanishing and exploding gradients in deep neural networks. It achieves this by carefully scaling the initial weights based on the size of the layers.

The key idea behind Xavier Initialization is to initialize the weights such that the signal propagating through the network neither diminishes nor grows excessively as it moves from one layer to the next. For a weight matrix $\mathbf{W} \in \mathbb{R}^{n_{in} \times n_{out}}$, where $n_{in}$ and $n_{out}$ are the number of input and output units, respectively, the weights are sampled from a uniform or a normal distribution.

For the uniform distribution the weights are initialized as:

$$W \sim \mathcal{U}\left(-\sqrt{\frac{6}{n_{in} + n_{out}}}, \sqrt{\frac{6}{n_{in} + n_{out}}}\right). \tag{8}$$

For the normal distribution the weights are initialized as:

$$W \sim \mathcal{N}\left(0, \frac{2}{n_{in} + n_{out}}\right). \tag{9}$$

This initialization strategy ensures that the variance of the inputs and gradients remains approximately constant across layers, promoting stable gradient flow during backpropagation.

#### 2.3.2 Kaiming Initialization

Kaiming Initialization [2], also known as He Initialization, is a weight initialization technique specifically designed to address the challenges posed by activation functions with non-negative outputs, such as Rectified Linear Units (ReLU) [25]. Unlike Xavier Initialization, which assumes symmetric activation functions such as sigmoid or tanh, Kaiming Initialization accounts for the asymmetric nature of ReLU and its variants, ensuring stable gradient flow during training.

The key idea behind Kaiming Initialization is to scale the initial weights based on the number of input units in the layer, taking into account the properties of ReLU activations. For a weight matrix $\mathbf{W} \in \mathbb{R}^{n_{in} \times n_{out}}$, where $n_{in}$ and $n_{out}$ are the number of input and output units, respectively, the weights are sampled from a uniform or a normal distribution.

For the normal distribution, the weights are initialized as:

$$W \sim \mathcal{N}\left(0, \frac{2}{n_{in}}\right). \tag{10}$$

For the uniform distribution, the weights are initialized as:

$$W \sim \mathcal{U}\left(-\sqrt{\frac{6}{n_{in}}}, \sqrt{\frac{6}{n_{in}}}\right). \tag{11}$$

This initialization strategy ensures that the variance of the outputs remains consistent across layers, preventing issues such as vanishing or exploding gradients in deep networks with ReLU activations. By explicitly considering the rectifying behavior of ReLU, Kaiming Initialization has become a cornerstone for training modern neural networks, particularly convolutional neural networks (CNNs) and other architectures that rely heavily on ReLU-based activations.

#### 2.3.3 Gaussian Initialization

Initializing weights from a Gaussian distribution has become the de facto standard in contemporary LLM literature. Unlike traditional initialization schemes such as Xavier or Kaiming, which were designed for specific activation functions and network structures, Gaussian Initialization relies on simpler assumptions about weight distributions. In this approach, weights are typically drawn from a normal distribution $\mathcal{N}(0, \sigma^2)$, where $\sigma$ is a constant standard deviation that is usually set to 0.02.

The widespread adoption of Gaussian Initialization in modern architectures can be attributed to the increasing complexity of these models. Transformers, for instance, employ intricate mechanisms such as self-attention, multi-head projections, and layer normalization, which often render traditional initialization schemes like Xavier or Kaiming less applicable. These architectures frequently utilize non-standard activation functions (e.g., GLU [26]) or entirely omit nonlinearities in certain layers (e.g., residual connections), making it challenging to derive optimal scaling factors based on activation properties.

Empirical evidence suggests that Gaussian Initialization performs well in practice, even without the explicit variance adjustments prescribed by Xavier or Kaiming. For example, many state-of-the-art Transformer implementations initialize weights using $\mathcal{N}(0, \sigma^2)$ and rely on subsequent normalization techniques such as LayerNorm or RMSNorm to stabilize activations during training [27, 28]. This reliance on normalization layers diminished the perceived importance of carefully calibrated initialization schemes, as the normalization process inherently mitigates issues related to improper scaling.

### 2.3.4 Special Residual Initialization

"Special Residual Initialization" [29] as proposed by the authors of GPT-2 is designed to stabilize the initial signal propagation through residual connections. They proposed to *only* scale down the initialization of the weights of the *last linear transformation* in each MLP and Attention layer by a factor of $\frac{1}{\sqrt{2N}}$, where $N$ is the number of layers in the network. Specifically, those weights are initialized using a normal distribution with scaled variance:

$$W \sim \mathcal{N}\left(0, \frac{\sigma^2}{2N}\right). \tag{12}$$

This ensures that the magnitude of the residual transformations remains small relative to the identity mapping at initialization, preventing the residual branches from dominating the forward pass and gradient flow. The special residual initialization strategy has proven particularly effective in stabilizing the training of GPT-2. However, to the best of our knowledge, there is no clear publicly available benchmark that shows whether this strategy also works well in larger LLM architectures.

### 2.3.5 Depth-Scaled Initialization (DS-Init)

DS-Init [13] scales the weights based on the layer index to control the variance of activations across layers. The authors observed that the large output variance of residual connections (RC) contributes to gradient vanishing issues. To mitigate this, they proposed to modify the standard weight initialization method. Specifically, they changed the original uniform distribution initialization from:

$$\mathbf{W} \in \mathbb{R}^{n_{\text{in}} \times n_{\text{out}}} \sim \mathcal{U}\left(-\beta, \beta\right), \tag{13}$$

to a depth-scaled version:

$$\mathbf{W} \in \mathbb{R}^{n_{\text{in}} \times n_{\text{out}}} \sim \mathcal{U}\left(-\frac{\beta \alpha}{\sqrt{l}}, \frac{\beta \alpha}{\sqrt{l}}\right), \tag{14}$$

where:

$$\beta = \sqrt{\frac{6}{n_{\text{in}} + n_{\text{out}}}}, \tag{15}$$

$\alpha$ is a hyperparameter in the range $[0, 1]$, and $l$ denotes the layer index. The introduction of $\alpha$ provides flexibility in adjusting the scale of the weights, allowing for fine-tuning based on the specific architecture and training regime.

### 2.3.6 Layer Index Rescaling (LIR)

While the DS-Init approach has shown promising results in stabilizing training for deep Transformers, we propose an adaptation of this idea to better suit our needs. More specifically, we apply it within the context of a normal distribution rather than a uniform distribution. Additionally, we simplify the scaling mechanism by eliminating the $\alpha$ hyperparameter, opting instead for a direct division by the square root of the layer index. This leads to the following initialization scheme:

$$\mathbf{W} \in \mathbb{R}^{n_{\text{in}} \times n_{\text{out}}} \sim \mathcal{N}\left(0, \frac{\sigma^2}{l}\right), \tag{16}$$

where $\sigma^2$ represents the initial variance of the weights, and $l$ is the layer index. In practice, this means the weight is scaled by $\sqrt{l}$ after being initialized from a Gaussian distribution of standard deviation $\sigma$.

By adopting a normal distribution, we align with common practices in modern neural network initialization while maintaining the benefits of depth-scaled initialization. The removal of the $\alpha$ hyperparameter simplifies the implementation and reduces the need for additional tuning, making our approach more straightforward and potentially more robust. This adaptation ensures that the variance of the weights decreases as the layer depth increases, promoting more stable gradient flow.

### 2.4 Weight Rescaling During Training

#### 2.4.1 Unit Weight Rescaling (UWR)

Unit Weight Rescaling [9] controls the weight norm by explicitly rescaling it to the unit norm after a specified number of optimization steps for each layer. Mathematically, this is expressed as:

$$W \leftarrow \frac{W}{\|W\|}. \tag{17}$$

By enforcing the weights to have a unit norm, UWR prevents large increases in the gradient magnitude, thereby mitigating the risk of overfitting. At the same time, it ensures a sufficiently large effective learning rate, which is crucial for generalization performance. This approach offers a simple yet effective way to regularize the weight norm without the need for extensive hyperparameter tuning.

The original paper demonstrated the effectiveness and robustness of UWR across a variety of computer vision applications, including image classification, object detection, semantic segmentation, and crowd counting. Results showed that UWR outperformed traditional weight decay, implicit weight rescaling (weight standardization), and gradient projection (AdamP) in terms of both performance and stability. This method is proposed as an extension to batch normalization.

#### 2.4.2 Z-Score Weight Rescaling (ZWR)

In Z-Score Weight Rescaling [10] the weights are adjusted to maintain their initial variance after each epoch. Specifically, the method ensures that:

$$\text{Var}(y_e^l) \approx \text{Var}(y_{e+1}^l), \tag{18}$$

for epoch index $e$, layer index $l$, and layer output activation $y$. By enforcing this constraint, ZWR helps stabilize the training dynamics and prevents the variance from diverging uncontrollably.

The procedure consists of two main steps:

1. **Standardization**: In the first step, the weights are standardized by calculating the z-scores from the weights $W^l$ of a given layer $l$. This involves subtracting the mean and dividing by the standard deviation of the weights:

$$z^l \leftarrow \frac{W^l - \mu(W^l)}{\sigma(W^l)}. \tag{19}$$

Here, $\mu(W^l)$ and $\sigma(W^l)$ represent the mean and standard deviation of the weights in layer $l$, respectively.

2. **Rescaling**: In the second step, the z-scores $z^l$ are multiplied by the standard deviation $\sigma_{\text{init}}^l$ given by the initialization strategy. If the mean of the weights deviates from zero, the z-score changes the signs of values close to the mean value. To counteract this, the initial mean of the weights $\mu(W^l)$ has to be added again:

$$W^l \leftarrow z^l \cdot \sigma_{\text{init}}^l + \mu(W^l). \tag{20}$$

The authors of the original paper showed that ZWR effectively readjust the weights with a better internal structure while preserving the learned directions through experiments on small fully-connected neural network and multiple computer vision models. This approach allows the model to retain the structural information encoded in the weight orientations while resetting their magnitudes to more appropriate levels. One key advantage of ZWR is that it only relies on the weights of a layer and does not incorporate activation or gradient information. This makes it applicable at arbitrary moments during training, providing flexibility in its application.

#### 2.4.3 Target Variance Rescaling (TVR)

While the ZWR method provides a robust framework to maintain weight variance by enforcing its conditions at the end of each epoch, we propose an adaptation that applies the rescaling formula *within the same epoch* and targets custom *target standard deviation values* $\sigma^l_{\text{target}}$. This modification allows for more flexible variance management during training. The updated formula is as follows:

$$W^l \leftarrow z^l \cdot \sigma^l_{\text{target}} + \mu(W^l), \tag{21}$$

By applying the rescaling formula within the same epoch, our approach ensures that weight variance is continuously managed, preventing unbounded growth and plasticity loss during training. This is particularly important for LLMs, where extended training regimes and immense parameter spaces make it challenging to maintain stable training dynamics. Furthermore, LLM pretraining is usually done only for one epoch.

## 3 Experiments

### 3.1 Setup

In this section, we describe the experimental setup used to evaluate our proposed approach. The experiments were conducted using a standard 1-billion-parameter LLaMA architecture [30] as implemented in Hugging Face Transformers [27] version 4.47.1. The training data consists of the SmolLM [31] dataset, which combines three diverse sources: FineWebEdu-Deduplicated, Cosmopedia-V2 and Python-Edu; we randomly sample 100 billion (100BT) from this corpus. The context length is fixed at 2048 tokens per sequence with packing mechanism. We did *not* use a dedicated attention mask that prevents self-attention between different documents within the same sequence, since it has limited impact on pre-training with standard context length [30]. All experiments were conducted using the Distributed Data Parallelism (DDP) strategy on 3 nodes, each equipped with 8 NVIDIA H100 GPUs. See Table 1 for training hyperparameters. The exact model configuration and tokenizer details are presented in Appendix 6.1.

| Hyperparameter | Value |
| --- | --- |
| Optimizer | Fused AdamW ($\beta_1 = 0.9$, $\beta_2 = 0.95$, $\epsilon = 1 \times 10^{-7}$) |
| Learning Rate Schedule | Linear warmup followed by cosine decay |
| Max Learning Rate | $6 \times 10^{-4}$ |
| End Learning Rate | $6 \times 10^{-5}$ |
| Warmup Tokens | 1 billion tokens (1BT) |
| Weight Decay | 0.1 (AdamW implementation) |
| Global Batch Size | 4032 |
| Precision | Mixed Precision BFloat16 |

Table 1: Default training hyperparameters for the presented pre-training experiments (unless specified otherwise).

### 3.2 Evaluation

**Metrics Monitored**   We tracked the following metrics to evaluate both the training dynamics and the overall effectiveness of our variance control strategies:

- **Training Loss**: To measure the convergence and stability of the model during training.

- **Downstream Task Performance**: To evaluate the model's generalization capabilities on a diverse set of tasks. We used the LM-Eval framework [32] as a standardized and reproducible approach to benchmarking large language models across a wide range of tasks, ensuring a fair comparison of model capabilities.

- **Throughput**: To assess the computational efficiency of the training process, ensuring that our methods do not introduce significant overhead.

- **Evolution of Standard Deviation**: To monitor how the standard deviation of weights evolves over time, providing insights into the effectiveness of our variance control techniques.

- **Extreme Activation Values**: To identify potential issues such as massive activations [33], which can lead to numerical instability or challenges during quantization or low-precision training.

**Downstream Tasks**   We evaluated our model on a diverse set of tasks designed to measure various aspects of language understanding and reasoning, including common sense reasoning tasks (HellaSwag [34], PIQA [35], SIQA [36], WinoGrande [37]), question answering (TriviaQA [38]), and multi-step reasoning tasks (ARC-Challenge (ARC-C), ARC-Easy (ARC-E) [39]).

### 3.3   Variance Control at Initialization

Effective variance control at initialization is crucial to ensure stable gradient flow, and consequently to improve downstream task performance in LLMs. In this set of experiments, we explored two key aspects of initial variance control, namely the search for optimal initialization standard deviation values for basic Gaussian initialization, and subsequently the comparison of this optimized baseline against LIR-based weight initialization.

**Searching for Optimal Initialization Standard Deviation Values**   While Gaussian initialization is widely used in modern neural networks, there is limited literature on benchmarking different standard deviation values specifically for LLMs. A common heuristic is to use a fixed standard deviation of 0.02 for weight initialization. However, this value is not guaranteed to be universally optimal and may not account for the unique characteristics of different model sizes or architectures.

We benchmarked various initialization standard deviation values for 2D modules in the decoder layers of our LLaMA-1B model. Our goal was to determine whether alternative initialization values could lead to better generalization and downstream task performance. By systematically varying the standard deviation of the Gaussian distribution, we aimed to identify initialization schemes that strike an optimal balance between signal propagation and training stability.

**Applying LIR**   In addition to searching for optimal initialization values, we also investigated the impact of applying LIR. We benchmarked the effect of LIR in combination with different initialization standard deviation values to evaluate whether this technique could further enhance the stability and performance of the model during training.

### 3.4   Variance Control during Training

We applied TVR in tandem with LIR to further control the weight variance growth during training. A key aspect of our approach was the exploration of optimal target standard deviation values for TVR. We conducted experiments to identify the most effective values for $\sigma_{\text{target}}^{(\ell)}$, considering both fixed and layer-specific configurations. Our goal was to determine whether optimizing this hyperparameter could lead to improved generalization and downstream task performance.

## 4   Results

In this section, we present the results of our experiments to evaluate the effectiveness of our proposed variance control methods LIR and TVR in improving the training dynamics and downstream performance of a standard 1B LLaMA LLM. Our analysis focused on two primary aspects: variance control at initialization, and variance control during training. We observed that careful management of weight variance not only enhances model stability, but also leads to measurable improvements in generalization and downstream task performance. Additionally, we also show evidence indicating that our variance control methods reduce extreme activation values.

### 4.1   Controlled Variance at Initialization

**Searching for Optimal Initialization Standard Deviation Values**   Starting with Figure 1, we observed that the standard deviation of the weights grows during training, but remains *partially bounded* due to weight decay (see also ablation in Appendix 6.2.1). However, the weights still deviate significantly from their initial distribution, particularly for lower initialization values. Notably, the rate of increase in standard deviation is much faster when starting with smaller initial values, such as 0.006 or 0.008, compared to larger values such as 0.02. This rapid growth highlights the sensitivity of weight dynamics to initialization choices. Additionally, the reduction in standard deviation for initial value of 0.03 is also caused by the partial bounding effect of weight decay.

As shown in Table 2, initial values within the 0.006 and 0.015 range consistently outperformed the baseline of 0.02 on downstream benchmarks. Figure 2 also shows the consistently positive effect of initial values on performance within this range throughout training.

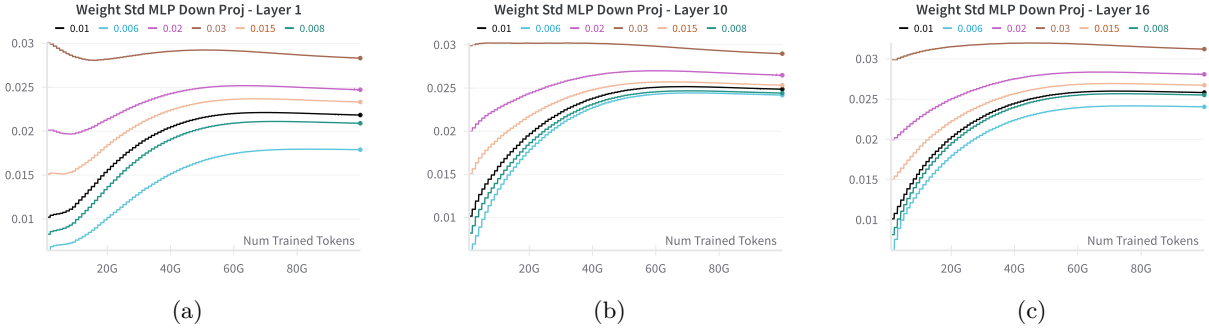

Figure 1: (a) Layer-1 (first); (b) Layer-10; (c) Layer-16 (last); standard deviation evolution of MLP Down Projection matrices for different initial standard deviation values $\sigma_{\text{init}}$. Note that the curves are not perfectly smooth due to the logging intervals of 1 billion tokens (1BT). More plots of standard deviation evolution are also presented in Appendix 6.3.1 and 6.3.2.

|  | Hella-Swag | PIQA | SIQA | WG | TQA | ARC-E | ARC-C | **Mean** | $\Delta_{\text{base}}$ |
|---|---|---|---|---|---|---|---|---|---|
| $\sigma_{\text{init}} = 0.03$ | 48.0 | 70.9 | 40.2 | 52.2 | 37.2 | 59.1 | 32.1 | 48.5 | -1.3 |
| $\sigma_{\text{init}} = 0.02$ (baseline) | 50.5 | 71.6 | 41.1 | 52.9 | 36.8 | 61.6 | 34.4 | 49.8 | 0.0 |
| $\sigma_{\text{init}} = 0.015$ | 51.6 | 71.6 | 40.4 | 52.8 | 37.3 | 60.4 | 33.7 | 49.7 | -0.1 |
| $\sigma_{\text{init}} = 0.01$ | 51.8 | 71.6 | 41.2 | 53.0 | 37.7 | **62.8** | 33.7 | 50.3 | 0.4 |
| $\sigma_{\text{init}} = 0.008$ | **51.9** | **71.8** | 41.0 | **54.5** | 35.6 | 62.4 | **36.0** | **50.5** | **0.6** |
| $\sigma_{\text{init}} = 0.006$ | 51.4 | 71.2 | **41.4** | 53.9 | **37.8** | 61.9 | 32.8 | 50.1 | 0.2 |

Table 2: Downstream Task Performance Benchmark Across Different Initialization Values.

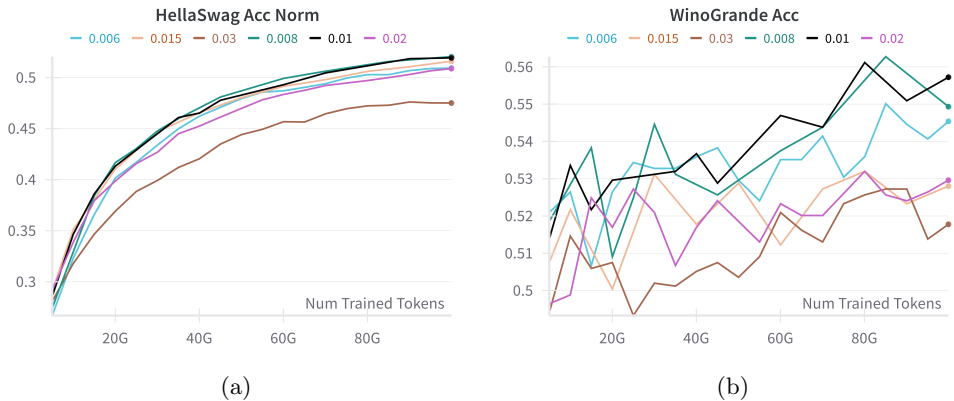

Figure 2: Evolution scores for (a) HellaSwag and (b) WinoGrande benchmarks during training for different initial standard deviation values $\sigma_{\text{init}}$.

**Applying LIR**   To further enhance these results, we explored the use of LIR. As illustrated in Figure 3, this technique accelerates the growth rate of standard deviation, particularly in deeper layers. Despite starting with lower initial standard deviation in deeper layers, they all ended up in a similar range compared to a scenario without LIR. Furthermore, the growth remained partially bounded due to the stabilizing effect of weight decay.

Figure 4a further underscores the benefits of LIR in consistently boosting downstream task performance for all initial standard deviation values. This finding is aligned with our hypothesis that controlling variance at initialization through layer-specific scaling can improve model generalization. Figure 4b and Figure 4c also shows the consistent effect of LIR throughout the training process.

For a more detailed analysis, additional ablation studies are provided in the Appendix. These include the impact of weight decay on the growth of weight standard deviation (see Appendix 6.2.1), as well as a comparison of LIR to the special residual initialization strategy introduced in GPT-2 (see Appendix 6.2.2). We furthermore present details on the evolution of standard deviation for all layers under different initialization values, both with and without LIR (see Appendix 6.3.1 and 6.3.2). Notably, we observed that the standard deviation of post-attention RMSNorm's scale vector $\gamma$ tends to increase with a *higher rate* for lower initial standard deviation values in the *early layers*, while it tends to increase with a *lower rate* in the *deeper layers* (Figure 23). We found a completely mirrored observation for the input RMSNorm, where the scale vector tends to increase with a *lower rate* for lower initial standard deviation values in the *early layers*, while it tends to increase with a *higher rate* in the *deeper layers* (see Figure 22).

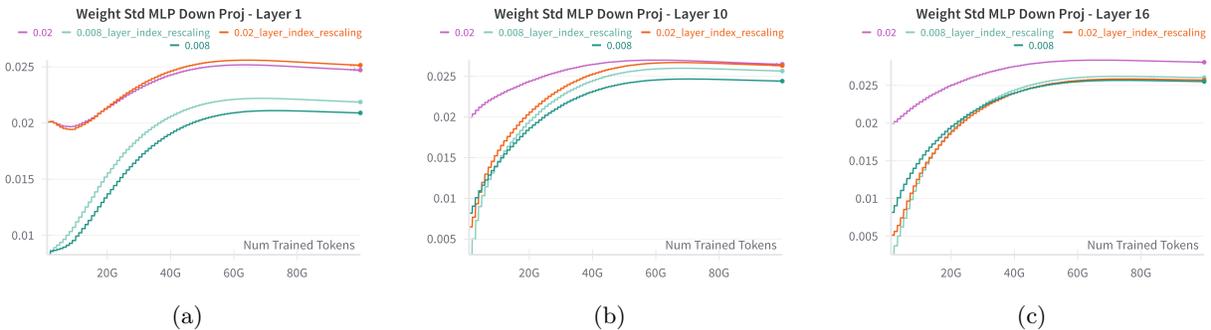

Figure 3: (a) Layer-1 (first); (b) Layer-10; (c) Layer-16 (last); comparison of standard deviation evolution of MLP Down Projection matrices for different initial standard deviation values $\sigma_{\text{init}}$ with and without LIR.

## 4.2 Controlled Variance during Training

To further stabilize the training by bounding the weight growth, we evaluated adding TVR to the aforementioned scenarios. We used an initial standard deviation value $\sigma_{\text{init}}$ of 0.006, combined with LIR. We applied TVR every 1 billion tokens (1BT) with a target standard deviation $\sigma_{\text{target}}$ of 0.01. These initial and target standard deviation values were chosen based on our ablation studies which can be seen in Appendix 6.2.3.

Our experiments demonstrated significant improvements in downstream task performance after applying TVR during pre-training (see Table 3), with up to 4.6% increase in HellaSwag compared to the baseline ($\sigma_{\text{init}} = 0.02$). Figure 5b illustrates that TVR complements weight decay by bounding the growth of weight standard deviation during training. Additionally, we observed that TVR may initially lead to slightly lower downstream task performance, but it recovers quickly and often surpasses baseline performance in later training stages (see Figure 5c). Plots of standard deviation evolution across layers for different target $\sigma_{\text{target}}$ values are also presented in Appendix 6.3.3, 6.3.4, and 6.3.5 for initial $\sigma_{\text{init}}$ values of 0.02, 0.008, and 0.006, respectively.

|  | Hella-Swag | PIQA | SIQA | WG | TQA | ARC-E | ARC-C | **Mean** | $\Delta_{\text{base}}$ |
|---|---|---|---|---|---|---|---|---|---|
| $\sigma_{\text{init}} = 0.02$ (baseline) | 50.5 | 71.6 | 41.1 | 52.9 | 36.8 | 61.6 | 34.4 | 49.8 | 0.0 |
| $\sigma_{\text{init}} = 0.008$+LIR+TVR($\sigma_{\text{target}} = 0.008$) | 54.7 | 72.2 | 42.2 | 57.1 | 38.1 | 64.0 | 35.5 | 52.0 | 2.1 |
| $\sigma_{\text{init}} = 0.006$+LIR+TVR($\sigma_{\text{target}} = 0.006$) | 53.7 | 72.4 | **43.6** | 55.6 | **40.6** | 63.3 | 36.1 | 52.2 | 2.4 |
| $\sigma_{\text{init}} = 0.006$+LIR+TVR($\sigma_{\text{target}} = 0.01$) | **55.1** | **73.4** | 42.7 | **57.0** | 38.9 | **64.3** | **36.1** | **52.5** | **2.7** |

Table 3: Downstream Task Performance Benchmark with LIR and TVR.

In summary, our experiments on a 1B LLaMA model demonstrate that controlling weight variance in pre-training–here specifically by carefully choosing $\sigma_{\text{init}}$, and applying LIR and TVR–plays a pivotal role in stabilizing the model and enhancing its downstream performance. Our findings indicate the importance of variance control as a foundational component of robust training methodologies, particularly for large-scale models. For readers interested in further details, comprehensive analyses of standard deviation evolution and ablation studies are provided in the Appendix, including the impact of varying the token interval for TVR and its relation to throughput (see Appendix 6.2.4), and a comparison of scaling $\sigma_{\text{target}}$ with depth versus keeping it constant (see Appendix 6.2.5).

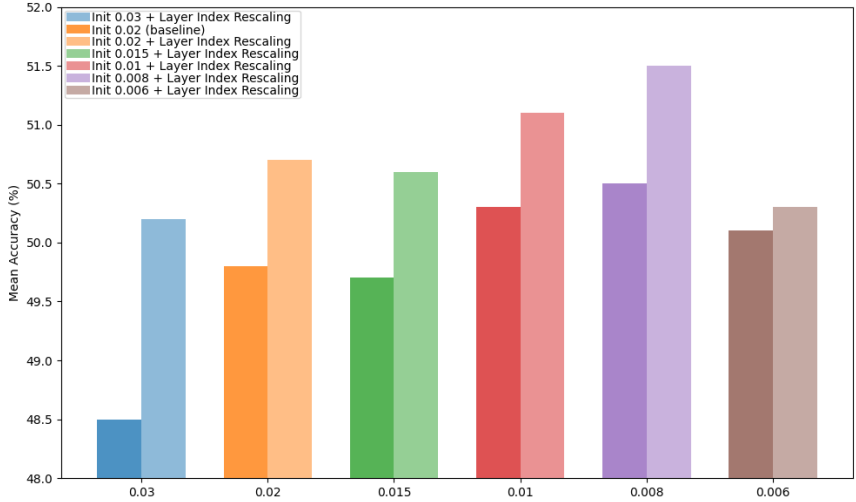

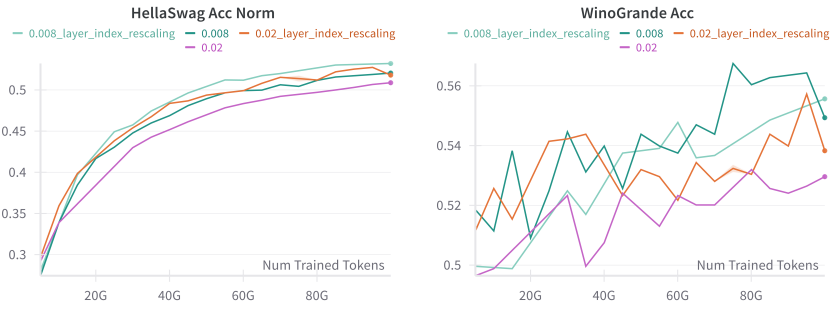

Figure 4: (a) Mean accuracy of all downstream task performance; (b) Evolution of HellaSwag scores; (c) Evolution of WinoGrande scores; comparison across different initial standard deviation values $\sigma_{\text{init}}$ with and without LIR.

### 4.3 Lower Extreme Activation Values

Massive activations in LLMs have recently emerged as a topic of interest[33, 40, 41]. A small subset of activations displays values that are orders of magnitude larger than the rest. These extreme activation values pose substantial risks, particularly for low-precision training and post-training quantization. All variance control techniques applied thus far have primarily aimed at managing the growth of weight variance. Given the tight coupling between weight variance and hidden state activation values, we conducted an analysis to assess how LIR and TVR influence activations, with a specific focus on these extreme activation values.

Figure 6 illustrates the comparison of maximum activation magnitude across layers for a given sentence "Summer is warm. Winter is cold."[1] between baseline and our best setup. Applying both LIR and TVR drastically reduced the extreme activation values. Further ablation studies on the relation between different initial and target standard deviation values towards extreme activation values are presented in 6.2.6.

---
[1]The same sentence is used in [33].

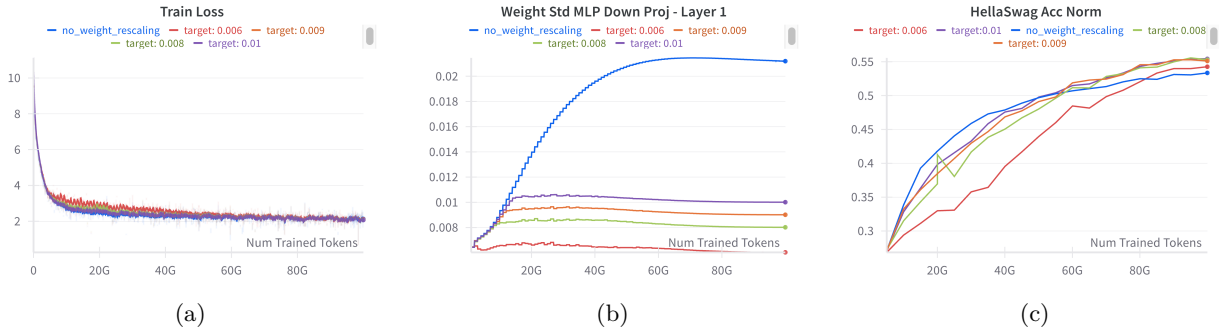

Figure 5: (a) Training loss curves; (b) Evolution of standard deviation of the MLP down projection in the first layer; (c) Evolution of HellaSwag scores; comparison for different TVR target standard deviation $\sigma_{\text{target}}$ with initial standard deviation $\sigma_{\text{init}} = 0.006$.

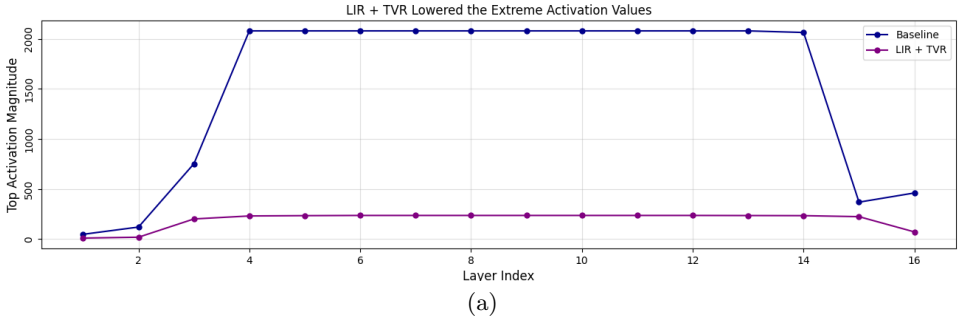

Figure 6: Comparison of maximum activation magnitude across layers for baseline ($\sigma_{\text{init}} = 0.02$) and LIR + TVR($\sigma_{\text{init}} = 0.006, \sigma_{\text{target}} = 0.01$) scenarios.

In conclusion, the combination of LIR and TVR seems to be highly effective in controlling weight variance and reducing extreme activation values across all layers. Together they seem to offer a practical solution to mitigate the risks posed by massive activations.

# 5 Conclusion and Future Work

In this work, we have explored the critical role of variance management in the training of LLMs, both at initialization time and during training. Our experiments demonstrate that carefully managing weight variance–through techniques such as optimal initialization standard deviation values, LIR and TVR–can significantly improve model stability, downstream task performance, and computational efficiency. Additionally, our approach reduces extreme activation values, mitigating risks associated with numerical instability in low-precision training and quantization.

While these results are promising, several avenues for future research remain:

**Scaling across Model Sizes, Architectural Variants and Training Compute Budgets** Our experiments were conducted using a 1-billion-parameter LLaMA model trained on 100 billion tokens. While these results are promising, it remains an open question whether the observed benefits of variance control will generalize to larger models and datasets. Training larger models with extended token counts introduces additional challenges, such as increased sensitivity to initialization and greater risks of unbounded weight growth. Assessing the scalability of our methods to such scenarios is a critical next step, and we plan to address this in future work within the scope of the compute resources available to us.

**Initialization Strategies for Embedding and LM Head Layers**   Our experiments to evaluate LIR and to determine optimal values for the initialization standard deviation were primarily focused on the decoder layers of the LLaMA model. However, we have yet to systematically explore dedicated initialization strategies for other critical components, such as the embedding layer and the language modeling (LM) head. These components play a crucial role in shaping the input and output representations of the model, and their initialization may significantly impact overall performance as well. Investigating optimal initialization schemes for these layers is an ongoing effort, and we plan to update our findings in future work.

**Extending Weight Rescaling to Normalization Layers**   While TVR has been successfully applied to the 2D weight matrices of decoder layers, its application to normalization layers–such as RMSNorm–remains underexplored. Specifically, the $\gamma$ vector in RMSNorm could potentially be adjusted in conjunction with the rescaling factor used for the 2D matrices. Exploring this interaction is a promising direction for future research.

**Potential for Higher Learning Rates**   One potential benefit of more stable training dynamics is the ability to use higher learning rates without compromising convergence or stability. Higher learning rates can accelerate training and improve computational efficiency, particularly for large-scale models. However, determining the optimal learning rate schedule in conjunction with variance control methods requires careful experimentation. This is an ongoing investigation, and we aim to provide more insights into this aspect in future updates.

# 6  Appendix

## 6.1  Model and Tokenizer

Table 4 shows the details of model architecture used in our experiment setup. The tokenizer used in this work is a frequency-truncated version of the Llama3 tokenizer, which includes a vocabulary size of 65,536 tokens.

| Parameter | Value |
| --- | --- |
| Hidden Size | 2048 |
| Intermediate Size | 5440 |
| Number of Hidden Layers | 16 |
| Number of Attention Heads | 16 |
| Number of Key-Value Heads | 16 |
| Activation Function | SiLU (Swish) |
| RMSNorm Epsilon | $1 \times 10^{-5}$ |
| Vocabulary Size | 65,536 |
| Maximum Position Embeddings | 2048 |

Table 4: Default model configuration for the presented Llama-1B pre-training experiments (unless specified otherwise).

## 6.2  Ablation Studies

### 6.2.1  What is the impact of weight decay in terms of the growth of weight standard deviation?

In this ablation study, we investigated the impact of weight decay in terms of the growth of weight variance. Figure 7 illustrates that weight decay helps in partially bounding the weight variance as can be seen in growing standard deviation when weight decay is not used (set to 0.0). Additionally, we found that weight decay also helps in downstream task performance as can be seen in Table 5 and Figure 8b.

| | Hella-Swag | PIQA | SIQA | WG | TQA | ARC-E | ARC-C | **Mean** | $\Delta_{\text{base}}$ |
| --- | --- | --- | --- | --- | --- | --- | --- | --- | --- |
| $\sigma_{\text{init}} = 0.02$ (baseline) | **50.5** | **71.6** | **41.1** | **52.9** | 36.8 | **61.6** | **34.4** | **49.8** | **0.0** |
| $\sigma_{\text{init}} = 0.03$ | 48.0 | 70.9 | 40.2 | 52.2 | 37.2 | 59.1 | 32.1 | 48.5 | -1.3 |
| $\sigma_{\text{init}} = 0.03$ + No Weight Decay | 46.2 | 70.0 | 39.9 | 52.1 | **38.6** | 56.8 | 32.8 | 48.1 | -1.8 |

Table 5: Downstream Task Performance Benchmark With and Without Weight Decay

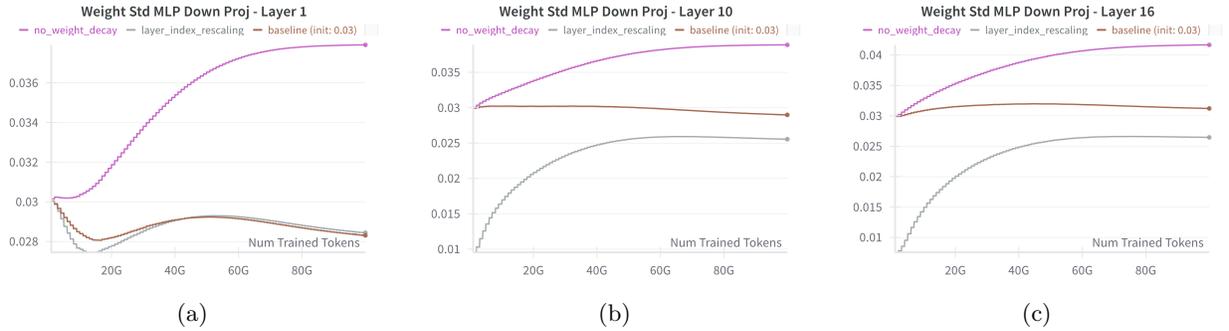

Figure 7: (a) Layer-1 (first); (b) Layer-10; (c) Layer-16 (last); comparison of standard deviation evolution of MLP Down Projection matrices with and without weight decay.

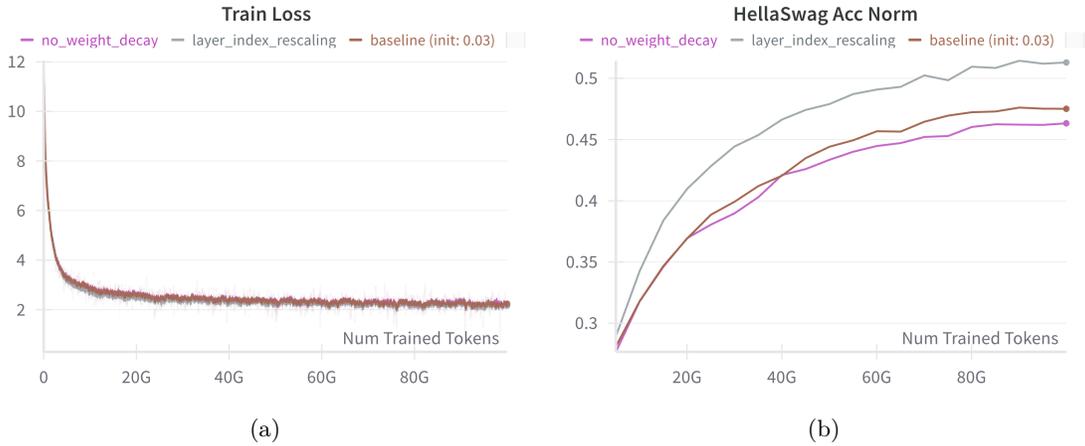

Figure 8: (a) Training loss curves; (b) Evolution of HellaSwag scores; with and without weight decay for initial std of 0.03.

### 6.2.2 Is the special initialization for residual weights introduced in GPT-2 better than layer index rescaling?

In this ablation study, we compare the effectiveness of LIR with the Special Init Residual (SIR) introduced in the GPT-2 paper. For this experiment, we used an initial standard deviation of 0.02 for weight initialization and did not apply TVR during training. Our findings indicate that while there is no significant difference in training loss between the two methods (Figure 9a), LIR demonstrates superior performance on downstream tasks compared to both the special initialization for residual weights and the baseline (Table 6). This suggests that LIR is a more effective strategy for improving generalization in large language models. Additionally, Figure 9b also shows that LIR consistently outperformed other methods throughout the training.

|  | Hella-Swag | PIQA | SIQA | WG | TQA | ARC-E | ARC-C | **Mean** | $\Delta_{\text{base}}$ |
|---|---|---|---|---|---|---|---|---|---|
| $\sigma_{\text{init}} = 0.02$ (baseline) | 50.5 | 71.6 | **41.1** | 52.9 | 36.8 | 61.6 | 34.4 | 49.8 | 0.0 |
| $\sigma_{\text{init}} = 0.02$ + SIR | 50.8 | 70.7 | 40.5 | 51.9 | **38.2** | **62.4** | 34.8 | 49.9 | 0.1 |
| $\sigma_{\text{init}} = 0.02$ + LIR | **52.5** | **72.0** | 40.5 | **54.6** | 37.8 | 62.3 | **35.2** | **50.7** | **0.9** |

Table 6: Downstream Task Performance Benchmark for Special Init Residual

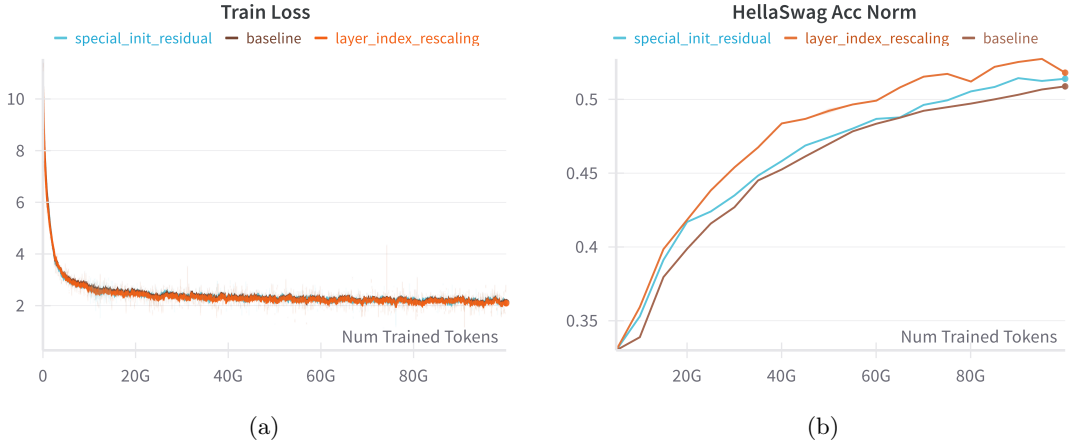

Figure 9: (a) Training loss curves; (b) Evolution of HellaSwag scores; comparison for LIR, special initialization for residual weights, and baseline.

### 6.2.3 Does weight rescaling help improve downstream task performance across varying initial and target standard deviation values?

The goal of this ablation study is to investigate the impact of TVR with different ($\sigma_{\text{init}}$) and ($\sigma_{\text{target}}$) values. Table 7 shows that, on average, $\sigma_{\text{init}}$ of 0.006 and $\sigma_{\text{target}}$ of 0.01 is the best combination. Figure 10 further shows the training loss curves comparison between $\sigma_{\text{init}} = 0.02$ and $\sigma_{\text{init}} = 0.006$. We argue that the observed cyclic loss behavior when using an initialization standard deviation $\sigma_{\text{init}} = 0.006$ and a $\sigma_{\text{target}} = 0.006$ can be attributed to the signal-to-noise ratio during training.

At smaller scales such as $\sigma_{\text{init}} = 0.006$, the relative magnitude of weight updates becomes disproportionately large compared to the initial weight values. This is particularly pronounced with adaptive optimizers like Adam, where the effective learning rate can amplify updates for smaller weights. As a result, the rescaling mechanism becomes overly reactive, frequently adjusting weights to maintain the target standard deviation. This heightened sensitivity disrupts the stability of the training process, contributing to cyclic loss patterns. Conversely, at $\sigma_{\text{init}} = 0.02$, the weight updates are proportionally smaller relative to the weight magnitudes, reducing the likelihood of abrupt changes and ensuring more stable training dynamics.

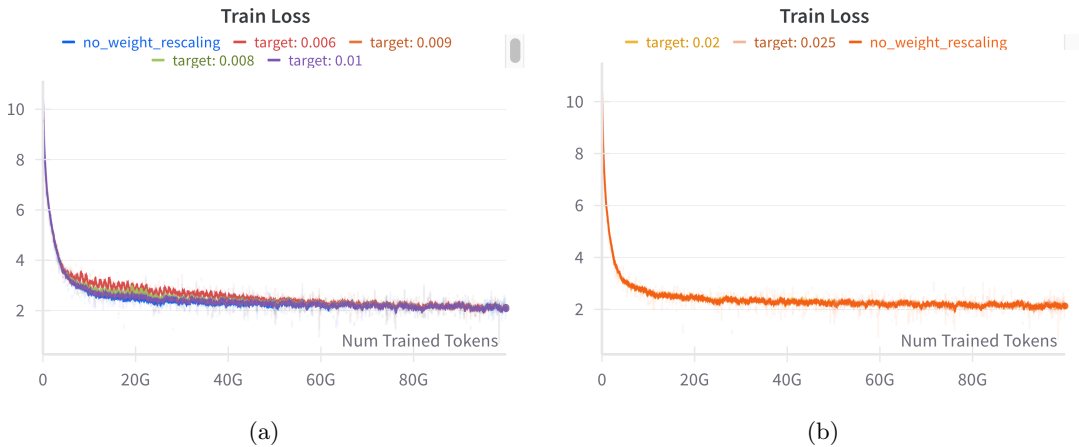

Figure 10: (a) $\sigma_{\text{init}} = 0.006$; (b) $\sigma_{\text{init}} = 0.02$; plots of training loss for different $\sigma_{\text{target}}$.

| | Hella-Swag | PIQA | SIQA | WG | TQA | ARC-E | ARC-C | **Mean** | $\Delta_{\text{base}}$ |
|---|---|---|---|---|---|---|---|---|---|
| $\sigma_{\text{init}} = 0.02$ (baseline) | 50.5 | 71.6 | 41.1 | 52.9 | 36.8 | 61.6 | 34.4 | 49.8 | 0.0 |
| + LIR | 52.5 | 72.0 | 40.5 | 54.6 | 37.8 | 62.3 | 35.2 | 50.7 | 0.9 |
| + LIR + TVR($\sigma_{\text{target}} = 0.02$) | 52.5 | 71.9 | 41.4 | 53.9 | 36.3 | 62.8 | 36.2 | 50.7 | 0.9 |
| + LIR + TVR($\sigma_{\text{target}} = 0.025$) | 52.5 | 72.4 | 41.4 | 53.0 | 36.1 | 63.0 | 34.6 | 50.4 | 0.6 |
| $\sigma_{\text{init}} = 0.015$ | 51.6 | 71.6 | 40.4 | 52.8 | 37.3 | 60.4 | 33.7 | 49.7 | -0.1 |
| + LIR | 52.2 | 71.8 | 41.9 | 53.6 | 38.2 | 61.5 | 34.7 | 50.6 | 0.7 |
| $\sigma_{\text{init}} = 0.01$ | 51.8 | 71.6 | 41.2 | 53.0 | 37.7 | 62.8 | 33.7 | 50.3 | 0.4 |
| + LIR | 53.3 | 71.3 | 42.0 | 55.6 | 37.9 | 62.2 | 35.7 | 51.1 | 1.3 |
| + LIR + TVR($\sigma_{\text{target}} = 0.01$) | 54.3 | 71.5 | 40.9 | **57.3** | 39.7 | 64.4 | **37.7** | 52.3 | 2.4 |
| + LIR + TVR($\sigma_{\text{target}} = 0.015$) | 54.0 | 71.9 | 41.6 | 55.7 | 37.5 | 62.5 | 35.6 | 51.3 | 1.4 |
| + LIR + TVR($\sigma_{\text{target}} = 0.02$) | 53.1 | 72.9 | 42.0 | 55.2 | 37.3 | 62.8 | 35.7 | 51.3 | 1.4 |
| + LIR + TVR($\sigma_{\text{target}} = 0.02$) | 52.2 | 71.8 | 41.9 | 53.6 | 38.2 | 61.5 | 34.7 | 50.6 | 0.7 |
| $\sigma_{\text{init}} = 0.008$ | 51.9 | 71.8 | 41.0 | 54.5 | 35.6 | 62.4 | 36.0 | 50.5 | 0.6 |
| + LIR | 52.4 | 72.1 | 41.3 | 56.0 | 38.0 | 64.3 | 36.1 | 51.5 | 1.6 |
| + LIR + TVR($\sigma_{\text{target}} = 0.008$) | 54.7 | 72.2 | 42.2 | 57.1 | 38.1 | 64.0 | 35.5 | 52.0 | 2.1 |
| + LIR + TVR($\sigma_{\text{target}} = 0.01$) | 55.1 | 71.5 | 41.6 | 54.2 | 37.6 | 63.6 | 37.1 | 51.5 | 1.7 |
| + LIR + TVR($\sigma_{\text{target}} = 0.015$) | 53.7 | 72.3 | 41.2 | 55.4 | 38.3 | 63.2 | 36.0 | 51.4 | 1.6 |
| + LIR + TVR($\sigma_{\text{target}} = 0.02$) | 53.0 | 71.4 | 42.2 | 57.2 | 36.6 | 64.1 | 37.1 | 51.7 | 1.8 |
| $\sigma_{\text{init}} = 0.006$ | 51.4 | 71.2 | 41.4 | 53.9 | 37.8 | 61.9 | 32.8 | 50.1 | 0.2 |
| + LIR | 52.5 | 71.2 | 40.1 | 53.6 | 39.2 | 61.8 | 33.9 | 50.3 | 0.5 |
| + LIR + TVR($\sigma_{\text{target}} = 0.006$) | 53.7 | 72.4 | **43.6** | 55.6 | 40.6 | 63.3 | 36.1 | 52.2 | 2.4 |
| + LIR + TVR($\sigma_{\text{target}} = 0.008$) | 54.6 | 71.3 | 41.2 | 55.6 | **43.5** | 64.3 | 35.7 | 52.3 | 2.5 |
| + LIR + TVR($\sigma_{\text{target}} = 0.009$) | 54.8 | 72.1 | 42.1 | 56.4 | 38.9 | **65.7** | 35.2 | 52.2 | 2.4 |
| + LIR + TVR($\sigma_{\text{target}} = 0.01$) | **55.1** | **73.4** | 42.7 | 57.0 | 38.9 | 64.3 | 36.1 | **52.5** | **2.7** |

Table 7: Downstream Task Performance Benchmark Across Different Initial and Target Standard Deviation Values

#### 6.2.4 What is the impact of changing weight rescaling frequency?

In this ablation study, we investigate the effect of varying the frequency of TVR. We used $\sigma_{\text{init}} = 0.006$ with LIR and set the target standard deviation $\sigma_{\text{target}} = 0.02$. Additionally, the $\sigma_{\text{target}}$ was not scaled with depth. We varied the interval (in tokens) at which TVR was applied and analyzed its impact on model performance and training efficiency. The evolution of the HellaSwag score over training is shown in Figure 11a, while the corresponding training throughput comparison is presented in Figure 11b. We can clearly see that there is no significant difference in terms of accuracy while higher tokens interval gave a slight jump in throughput.

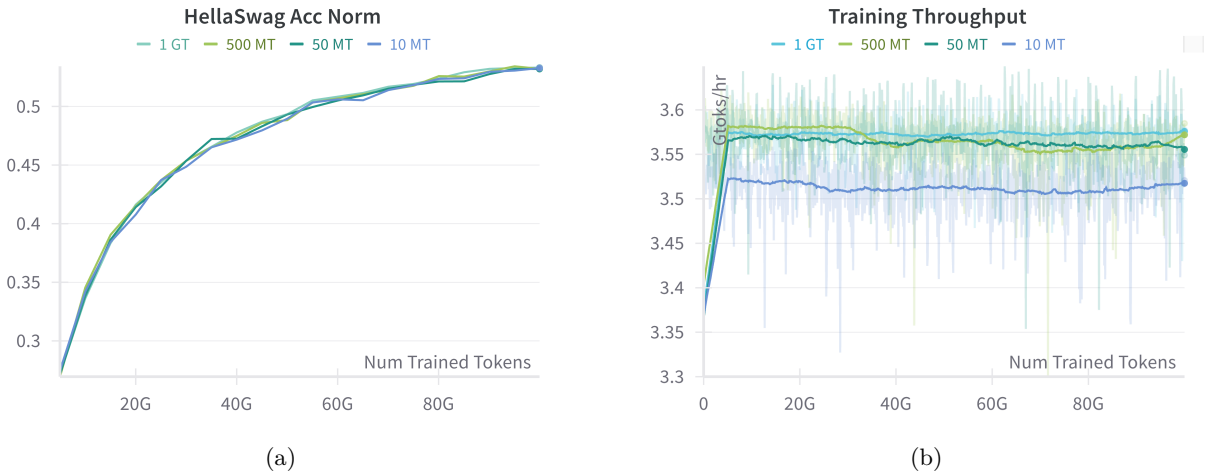

Figure 11: (a) Evolution of HellaSwag scores; (b) Training throughput; comparison for different TVR frequencies.

### 6.2.5  What is the impact of scaling the target standard deviation with depth?

In this ablation study, we investigate the effect of scaling the target standard deviation ($\sigma_{\text{target}}$) of TVR with depth. Specifically, we compare two strategies: (1) scaling $\sigma_{\text{target}}$ proportionally with depth using LIR, and (2) keeping $\sigma_{\text{target}}$ constant across all layers. For this experiment, we initialized the weights with a standard deviation of 0.006, applied LIR, and set the base target standard deviation for TVR to 0.01.

The results of the comparison are shown in Figure 12a (HellaSwag score evolution) and Figure 12b (training loss curves). Additionally, Figures 12c and 12d illustrate the evolution of the standard deviation for the MLP down-projection weights in the first and tenth layers, respectively.

Our findings reveal that scaling $\sigma_{\text{target}}$ with depth leads to cyclic fluctuations in the training loss, which adversely affects downstream task performance. In contrast, keeping $\sigma_{\text{target}}$ constant behaves similarly to introducing a "warmup period" for deeper layers. This is because the rate of increase in standard deviation is naturally faster for deeper layers compared to shallower ones-*Tenth layer reached 0.01 in around 8BT, while First layer only reached 0.01 in around 15BT*-as shown in Figures 12c and 12d. This gradual adjustment appears to stabilize training and improve generalization.

These results suggest that maintaining a fixed target standard deviation across layers is more effective than scaling it with depth. By avoiding abrupt changes in weight variance, this approach ensures smoother training dynamics and better performance on downstream tasks.

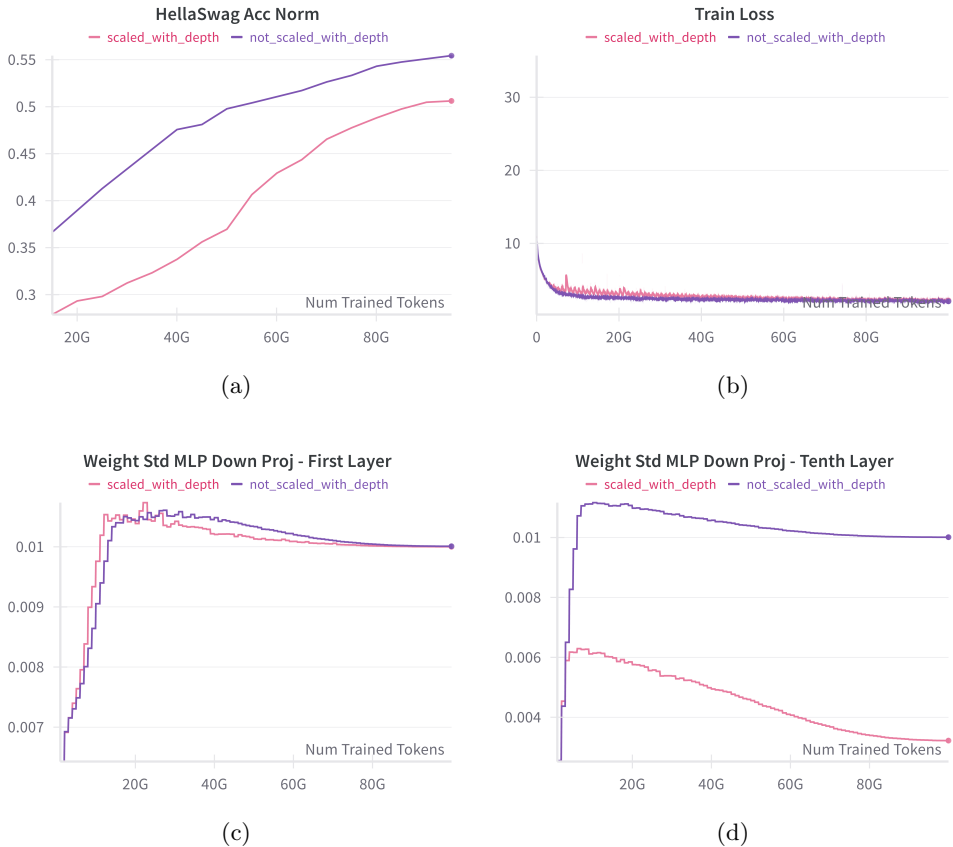

Figure 12: (a) Evolution of HellaSwag scores; (b) Training loss curves; (c) Evolution of standard deviation for the MLP down-projection weights in the first layer; (d) Evolution of standard deviation for the MLP down-projection weights in the tenth layer; comparison with and without scaling the target standard deviation with depth.

### 6.2.6 What is the relation between different initial and target standard deviation values towards extreme activation values?

In this ablation study, we analyzed what relation that $\sigma_\text{init}$ and $\sigma_\text{target}$ standard deviation values have in terms of extreme massive activation values. Figure 13 illustrates that there is a high positive correlation both between initial and/or target standard deviation values towards extreme activation values.

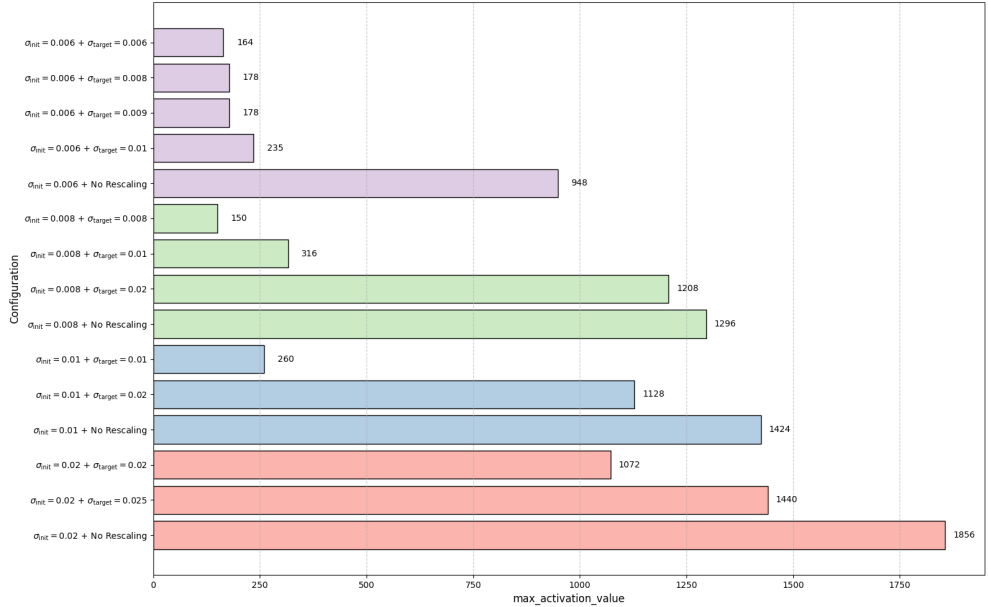

(a)

Figure 13: (a) Comparison of different $\sigma_\text{init}$ and $\sigma_\text{target}$ towards the maximum activation values across layers.

### 6.2.7 Is weight rescaling also useful for smaller, non-GLU-based LLMs?

In this ablation study, we investigated the impact of LIR and TVR on GPT-2-124M model. Table 9 shows the hyperparameters used in the experiment. The tokenizer used is default GPT-2 tokenizer. The context length is 1024. We found that LIR and TVR both boosted the downstream task performance as illustrated in Figure 14.

| Hyperparameter | Value |
| --- | --- |
| Optimizer | Fused AdamW ($\beta_1 = 0.9$, $\beta_2 = 0.95$, $\epsilon = 1 \times 10^{-7}$) |
| Learning Rate Schedule | Linear warmup followed by cosine decay |
| Max Learning Rate | $6 \times 10^{-4}$ |
| End Learning Rate | $6 \times 10^{-5}$ |
| Warmup Tokens | 1 billion tokens (1BT) |
| Weight Decay | 0.1 (AdamW implementation) |
| Global Batch Size | 2048 |
| Precision | Mixed Precision BFloat16 |
| Gradient Clipping | 1.0 |
| Dropout rate | 0.0 |
| Bias | not used |

Table 8: GPT-2 Training Hyperparameters

Table 9: Training hyperparameters for the presented GPT-2 pre-training experiments.

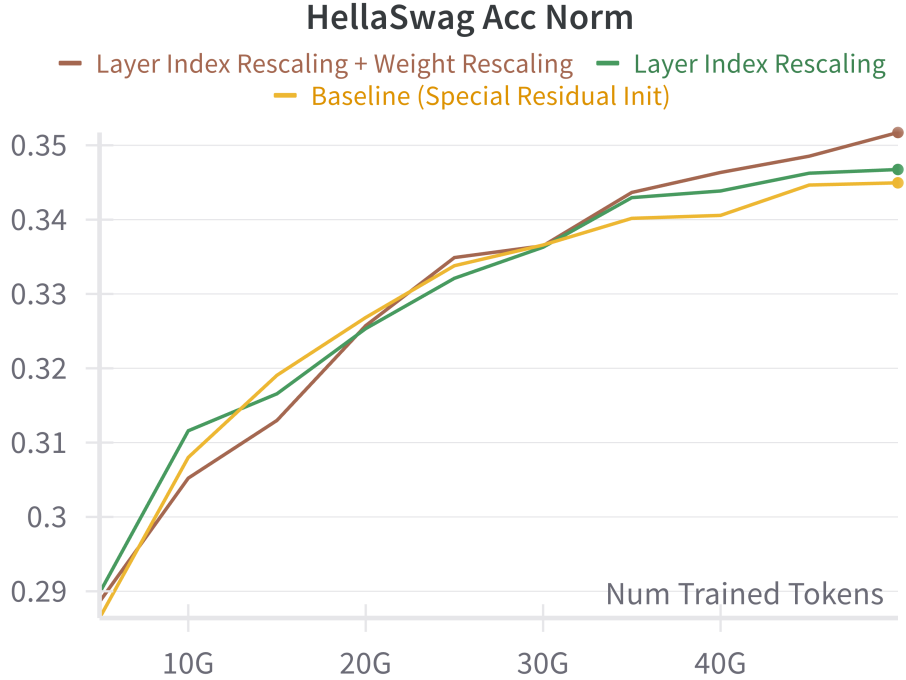

Figure 14: Evolution of HellaSwag scores comparing the baseline, LIR, and LIR + TVR.

## 6.3 Evolution of Standard Deviation Across Layers

### 6.3.1 Without Layer Index Rescaling and Without Weight Rescaling

Plots of standard deviation evolution of all 2D modules in the decoder layers of our Llama-1B models, including input RMSNorm, post-attention RMSNorm, embedding layer, and LMHead layer are presented in Figure 15 - Figure 24.

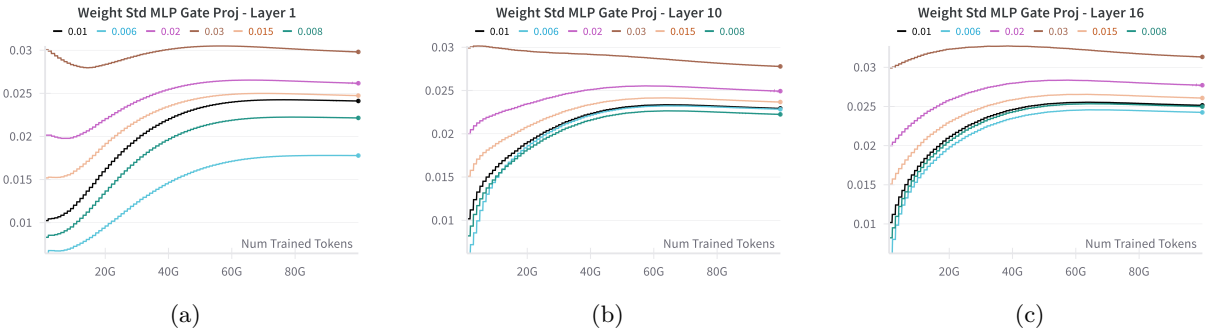

Figure 15: (a) Layer-1 (first); (b) Layer-10; (c) Layer-16 (last); standard deviation evolution of MLP Gate Projection matrices for different init standard deviation values.

### 6.3.2 With Layer Index Rescaling and Without Weight Rescaling

Plots of standard deviation evolution with applied LIR on all 2D modules in the decoder layers. The plots includes all the 2D modules in the decoder layers as well as the input RMSNorm, post-attention RMSNorm,

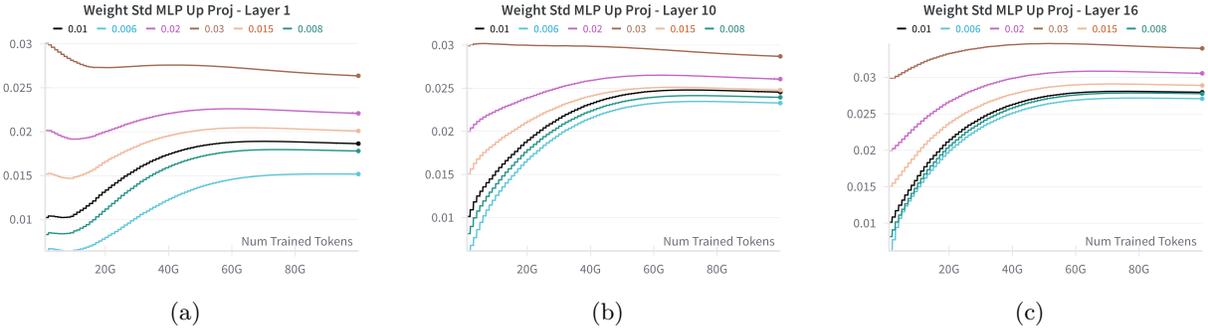

Figure 16: (a) Layer-1 (first); (b) Layer-10; (c) Layer-16 (last); standard deviation evolution of MLP Up Projection matrices for different init standard deviation values.

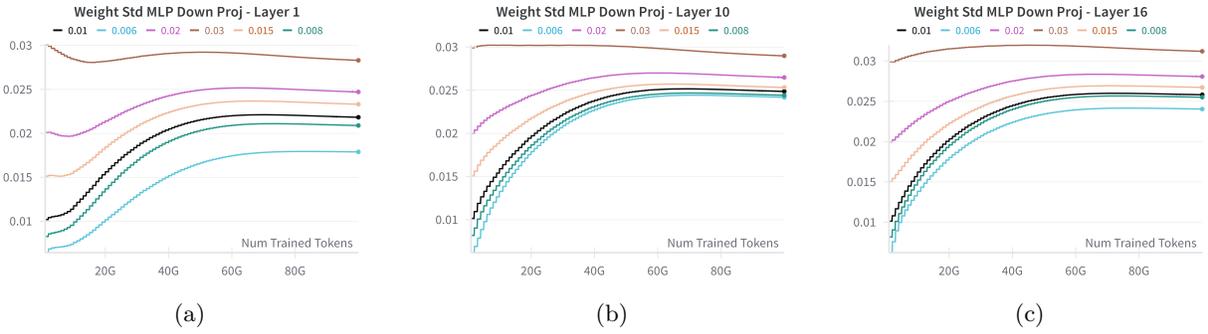

Figure 17: (a) Layer-1 (first); (b) Layer-10; (c) Layer-16 (last); standard deviation evolution of MLP Down Projection matrices for different init standard deviation values.

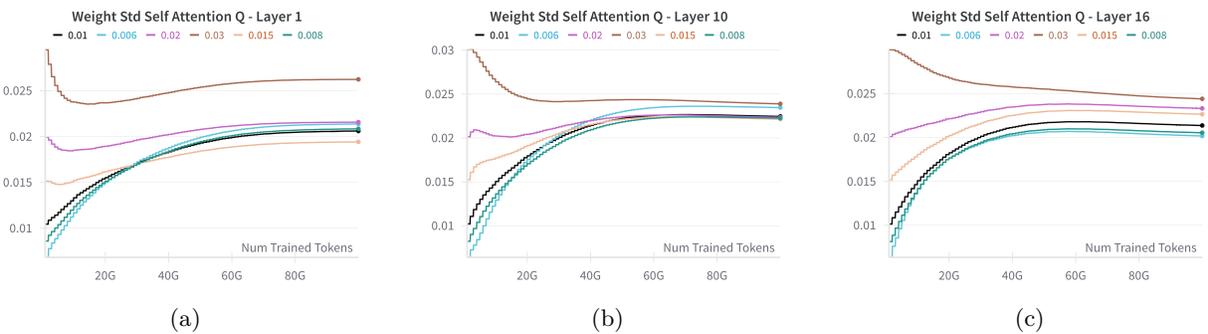

Figure 18: (a) Layer-1 (first); (b) Layer-10; (c) Layer-16 (last); standard deviation evolution of Attention Q projection matrices for different init standard deviation values.

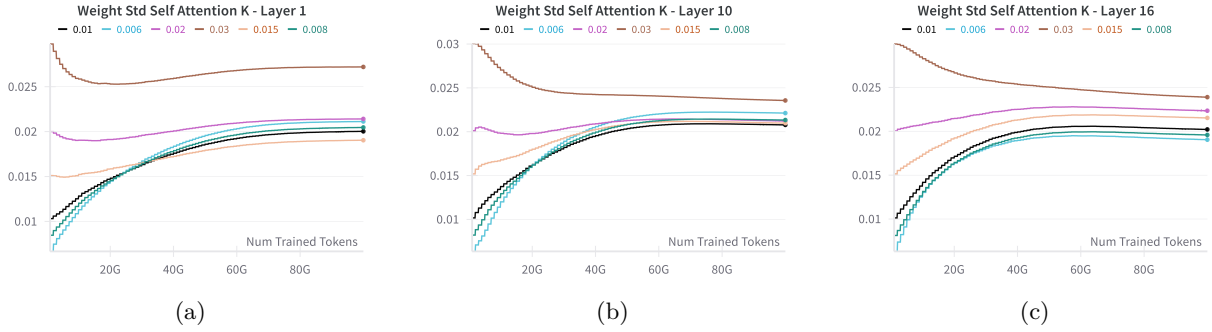

Figure 19: (a) Layer-1 (first); (b) Layer-10; (c) Layer-16 (last); standard deviation evolution of Attention K projection matrices for different init standard deviation values.

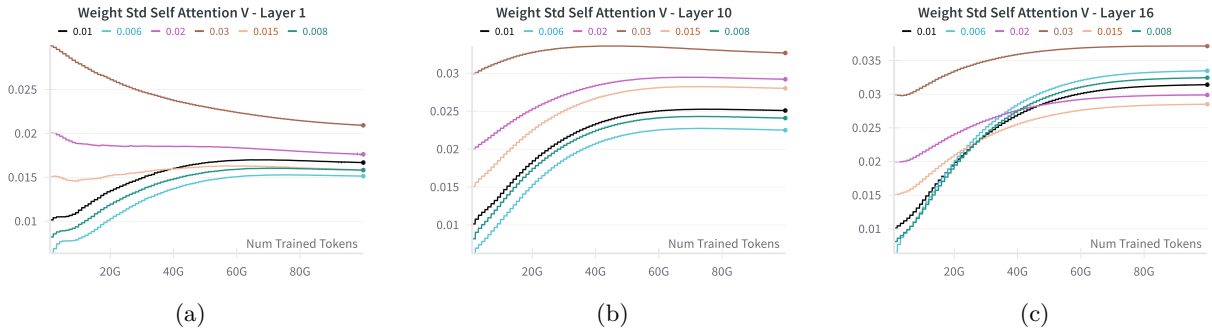

Figure 20: (a) Layer-1 (first); (b) Layer-10; (c) Layer-16 (last); standard deviation evolution of Attention V projection matrices for different init standard deviation values.

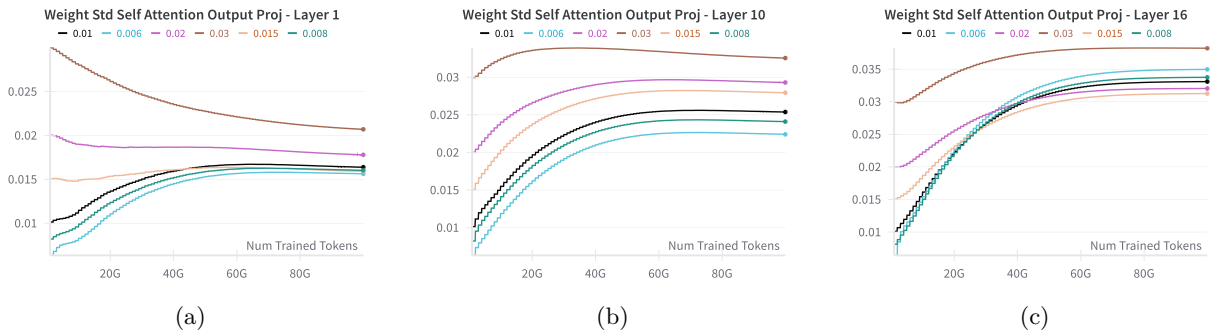

Figure 21: (a) Layer-1 (first); (b) Layer-10; (c) Layer-16 (last); standard deviation evolution of Attention Output projection matrices for different init standard deviation values.

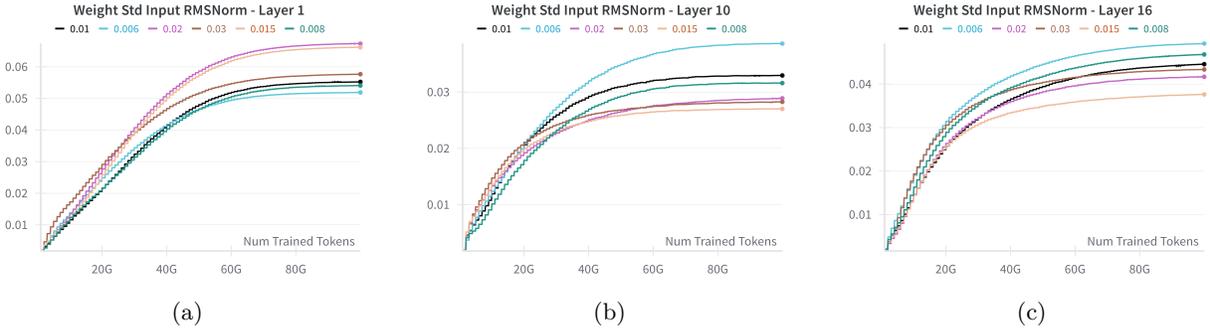

Figure 22: (a) Layer-1 (first); (b) Layer-10; (c) Layer-16 (last); standard deviation evolution of Input RMSNorm weight vectors for different init standard deviation values.

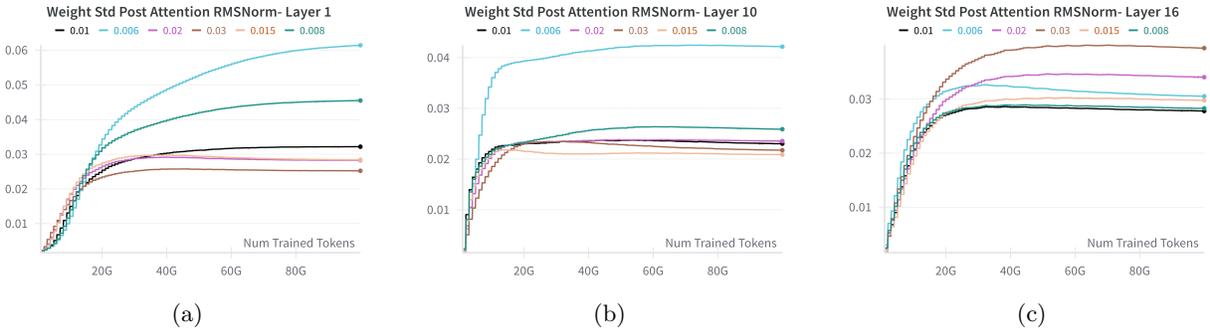

Figure 23: (a) Layer-1 (first); (b) Layer-10; (c) Layer-16 (last); standard deviation evolution of Post Attention RMSNorm weight vectors for different init standard deviation values.

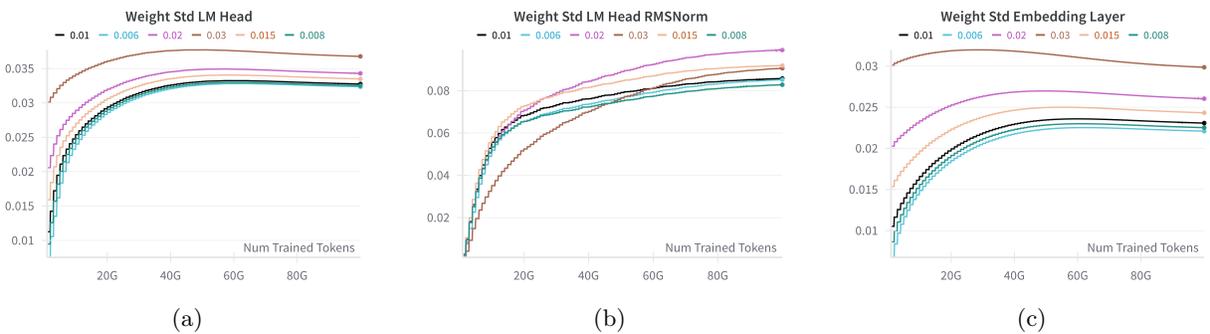

Figure 24: (a) LM-Head; (b) LM-Head RMSNorm; (c) Embedding Layer; standard deviation evolution for different init standard deviation values.

embedding layer, and LMHead layers. Colors in each plot represent different initial standard deviation values of the 2D modules in the decoder layers as presented in Figure 25 - Figure 34.

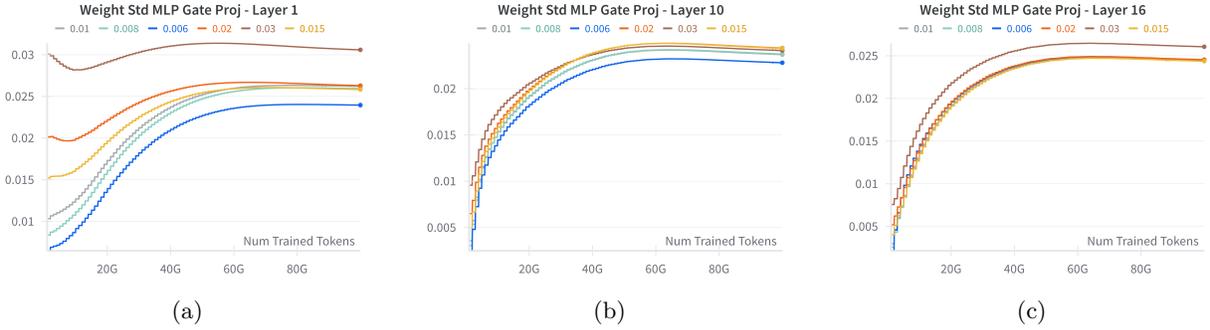

Figure 25: (a) Layer-1 (first); (b) Layer-10; (c) Layer-16 (last); standard deviation evolution (with LIR) of MLP Gate Projection matrices for different init standard deviation values.

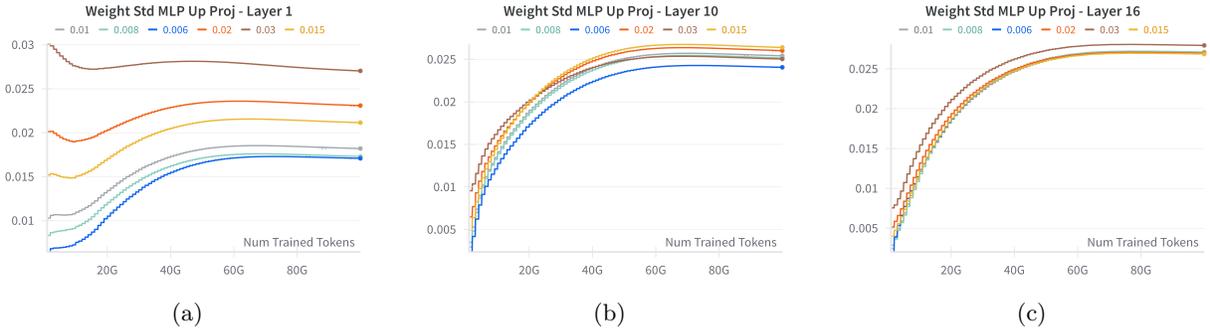

Figure 26: (a) Layer-1 (first); (b) Layer-10; (c) Layer-16 (last); standard deviation evolution (with LIR) of MLP Up Projection matrices for different init standard deviation values.

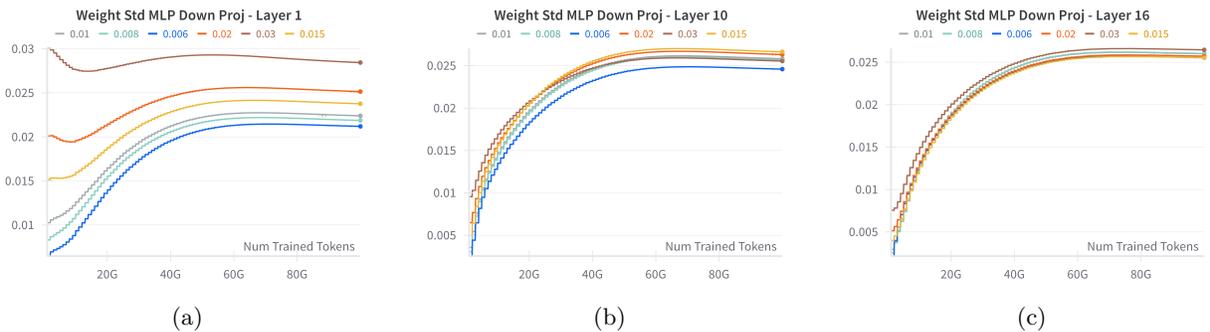

Figure 27: (a) Layer-1 (first); (b) Layer-10; (c) Layer-16 (last); standard deviation evolution (with LIR) of MLP Down Projection matrices for different init standard deviation values.

### 6.3.3 With Layer Index Rescaling and With Weight Rescaling (Initial Standard Deviation: 0.02

Plots of standard deviation evolution with applied LIR and TVR of all 2D modules in the decoder layers. The plots includes all the 2D modules in the decoder layers as well as the input RMSNorm, post-attention

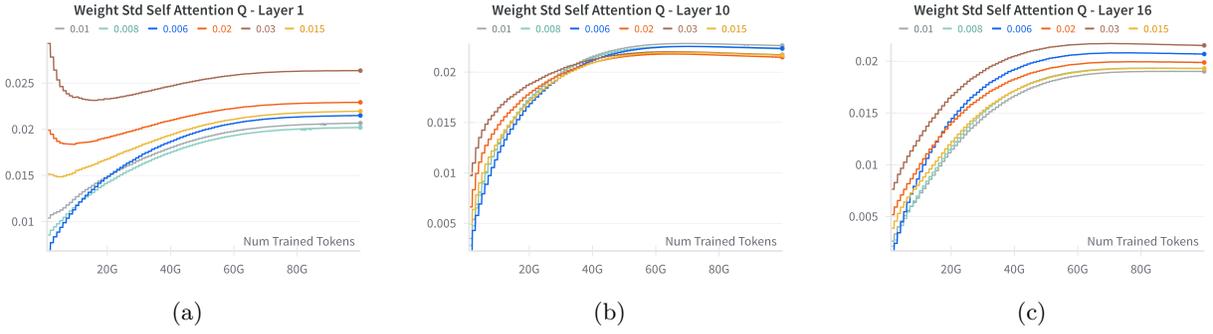

Figure 28: (a) Layer-1 (first); (b) Layer-10; (c) Layer-16 (last); standard deviation evolution (with LIR) of Attention Q projection matrices for different init standard deviation values.

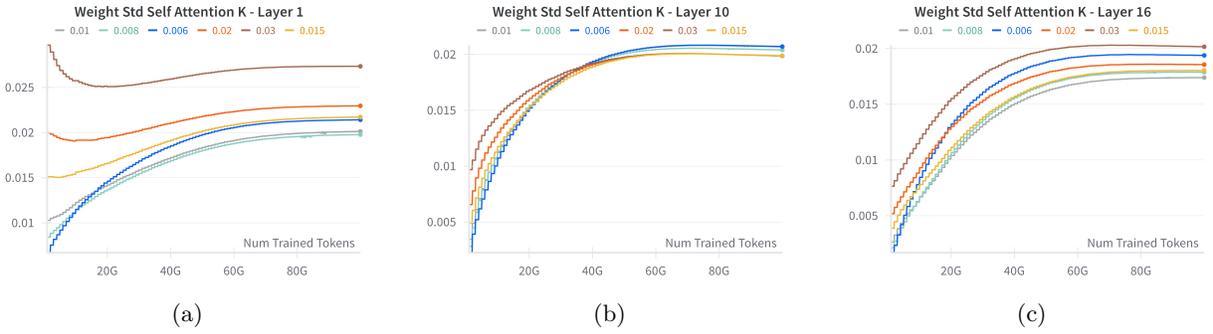

Figure 29: (a) Layer-1 (first); (b) Layer-10; (c) Layer-16 (last); standard deviation evolution (with LIR) of Attention K projection matrices for different init standard deviation values.

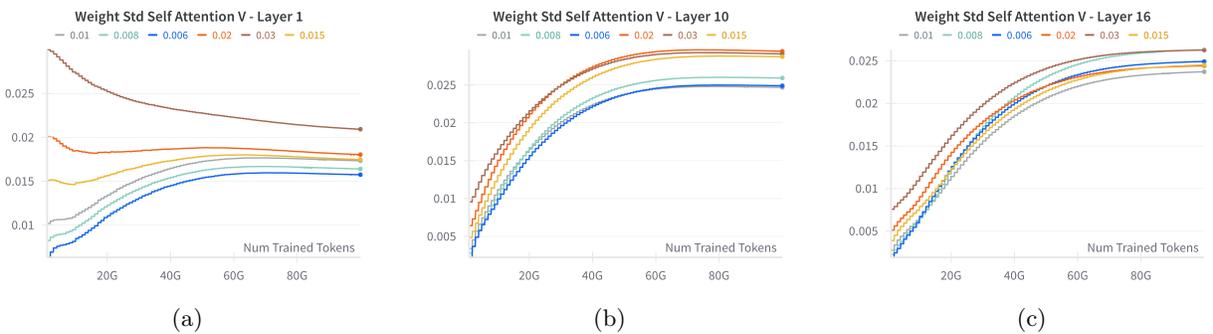

Figure 30: (a) Layer-1 (first); (b) Layer-10; (c) Layer-16 (last); standard deviation evolution (with LIR) of Attention V projection matrices for different init standard deviation values.

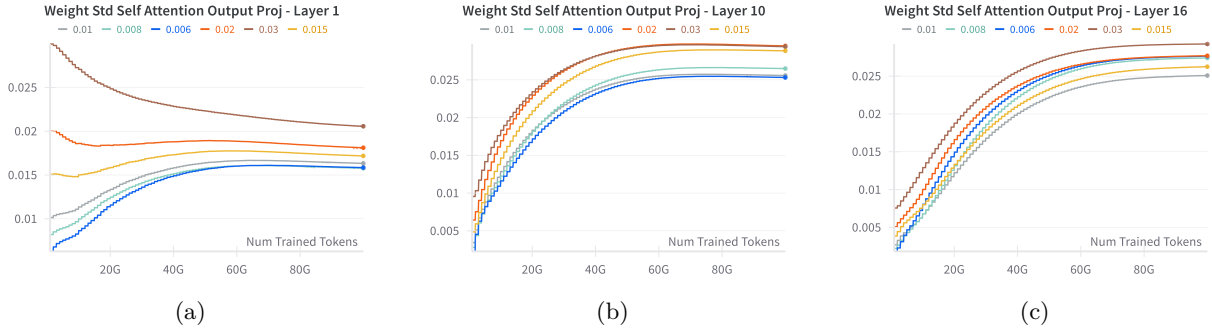

Figure 31: (a) Layer-1 (first); (b) Layer-10; (c) Layer-16 (last); standard deviation evolution (with LIR) of Attention Output projection matrices for different init standard deviation values.

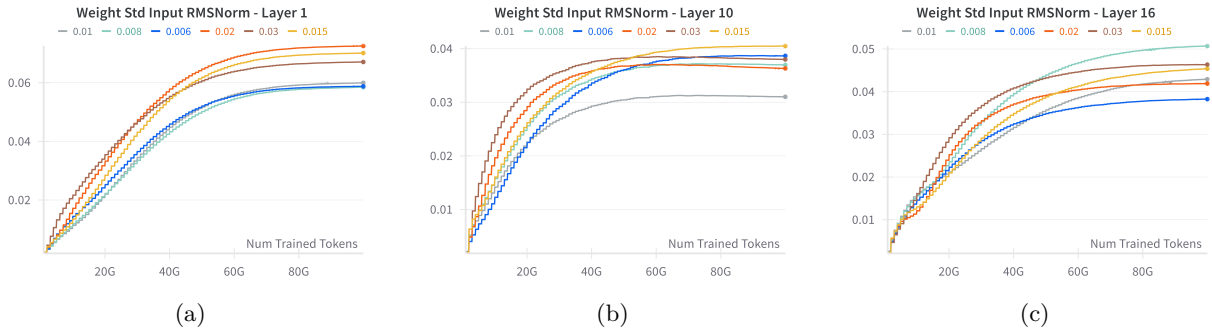

Figure 32: (a) Layer-1 (first); (b) Layer-10; (c) Layer-16 (last); standard deviation evolution (with LIR) of Input RMSNorm weight vectors for different init standard deviation values.

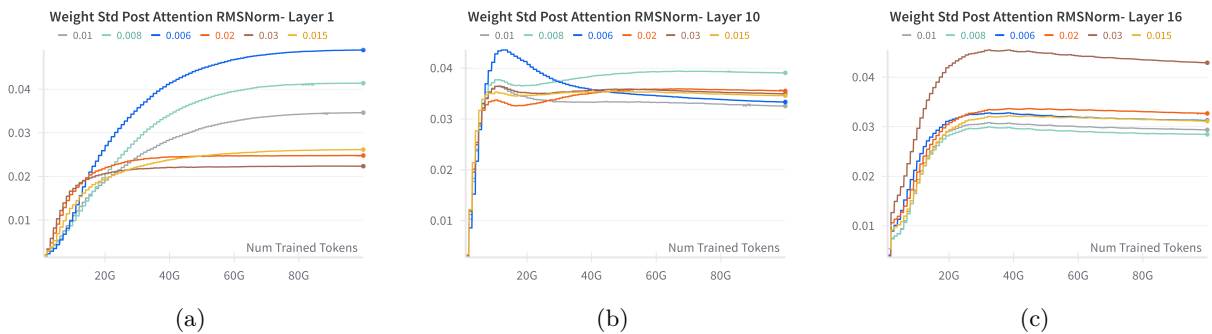

Figure 33: (a) Layer-1 (first); (b) Layer-10; (c) Layer-16 (last); standard deviation evolution (with LIR) of Post Attention RMSNorm weight vectors for different init standard deviation values.

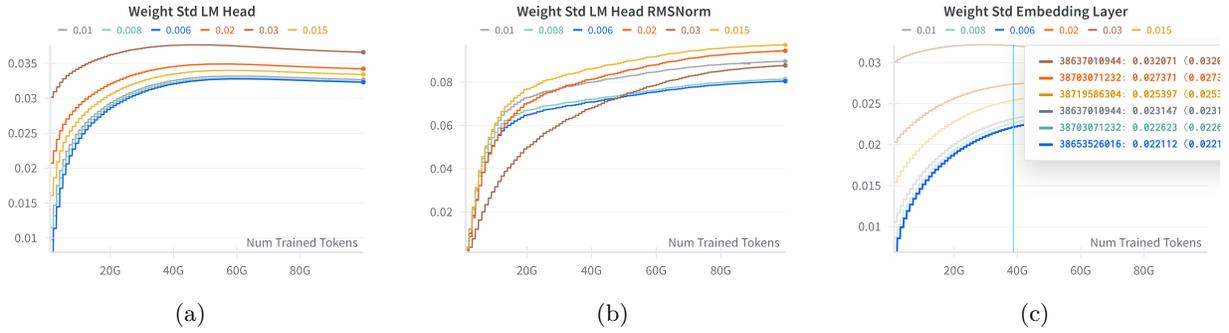

Figure 34: (a) LM-Head; (b) LM-Head RMSNorm; (c) Embedding Layer; standard deviation evolution (with LIR) for different init standard deviation values.

RMSNorm, embedding layer, and LMHead layers. Colors in each plot represent different target standard deviation values $\sigma_{\text{target}}$ of the TVR applied on 2D modules in the decoder layers as presented in Figure 35 - Figure 44.

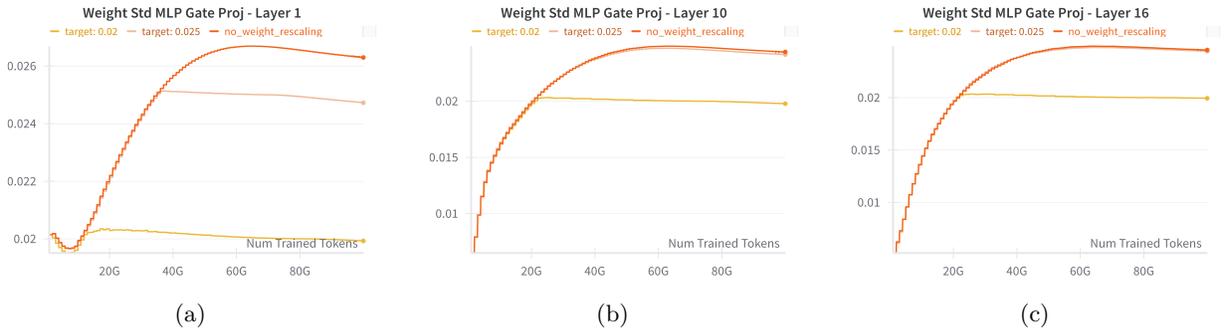

Figure 35: (a) Layer-1 (first); (b) Layer-10; (c) Layer-16 (last); standard deviation evolution (with LIR + TVR of init std 0.02) of MLP Gate Projection matrices for different init standard deviation values.

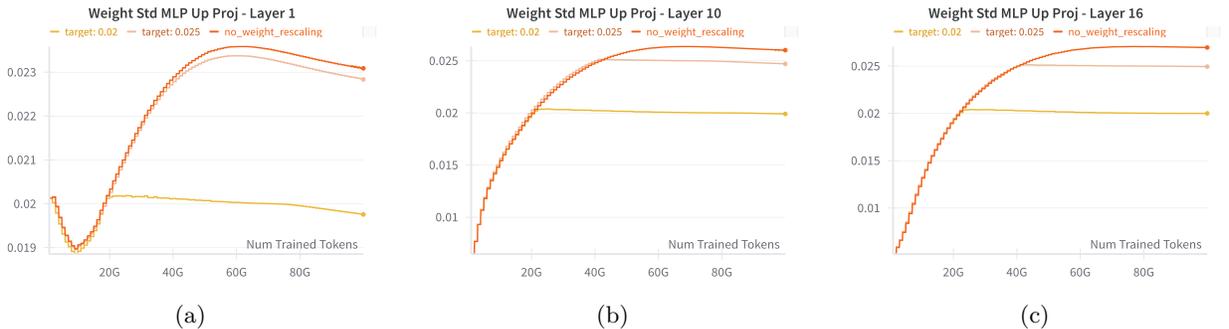

Figure 36: (a) Layer-1 (first); (b) Layer-10; (c) Layer-16 (last); standard deviation evolution (with LIR + TVR of init std 0.02) of MLP Up Projection matrices for different init standard deviation values.

### 6.3.4 With Layer Index Rescaling and With Weight Rescaling (Initial Standard Deviation: 0.008

Plots of standard deviation evolution with applied LIR and **TVR** of all 2D modules in the decoder layers. The plots includes all the 2D modules in the decoder layers as well as the input RMSNorm, post-attention

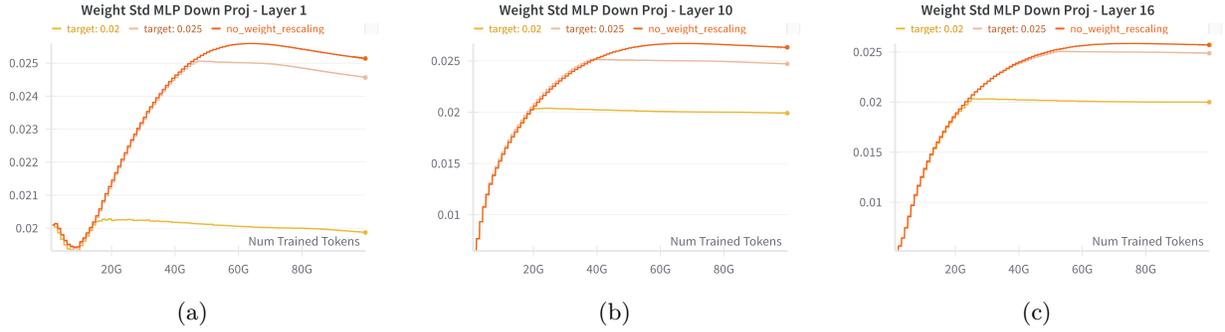

Figure 37: (a) Layer-1 (first); (b) Layer-10; (c) Layer-16 (last); standard deviation evolution (with LIR + TVR of init std 0.02) of MLP Down Projection matrices for different init standard deviation values.

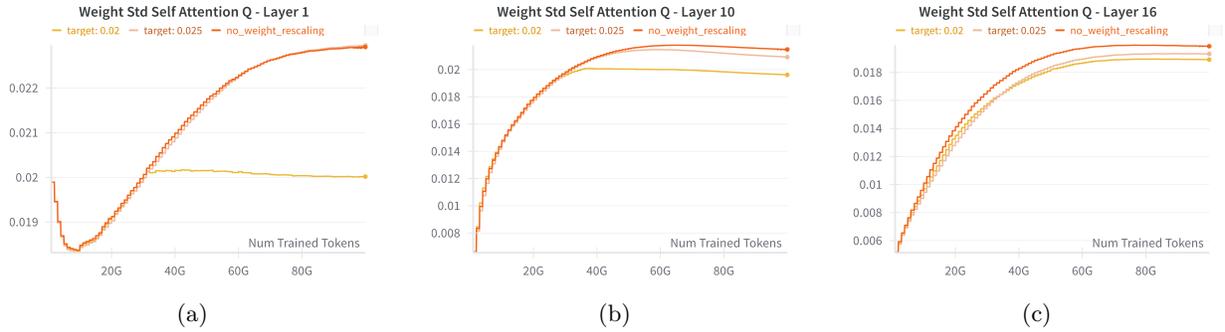

Figure 38: (a) Layer-1 (first); (b) Layer-10; (c) Layer-16 (last); standard deviation evolution (with LIR + TVR of init std 0.02) of Attention Q projection matrices for different init standard deviation values.

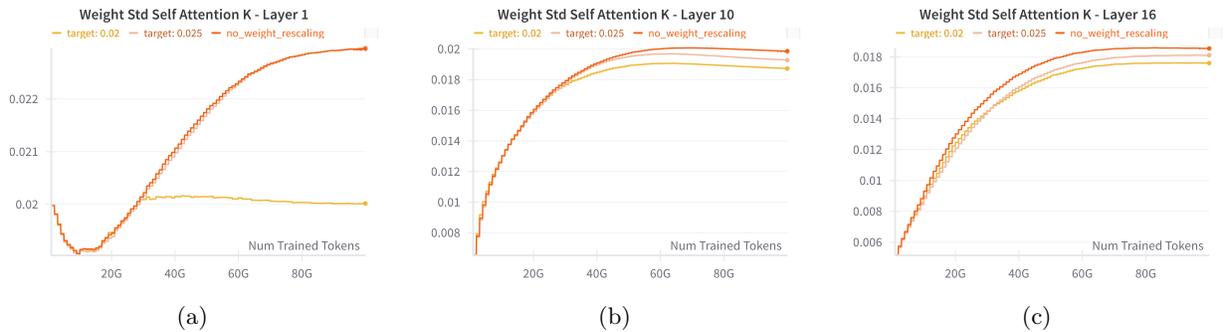

Figure 39: (a) Layer-1 (first); (b) Layer-10; (c) Layer-16 (last); standard deviation evolution (with LIR + TVR of init std 0.02) of Attention K projection matrices for different init standard deviation values.

Variance Control via *Weight Rescaling* in LLM Pre-training       A PREPRINT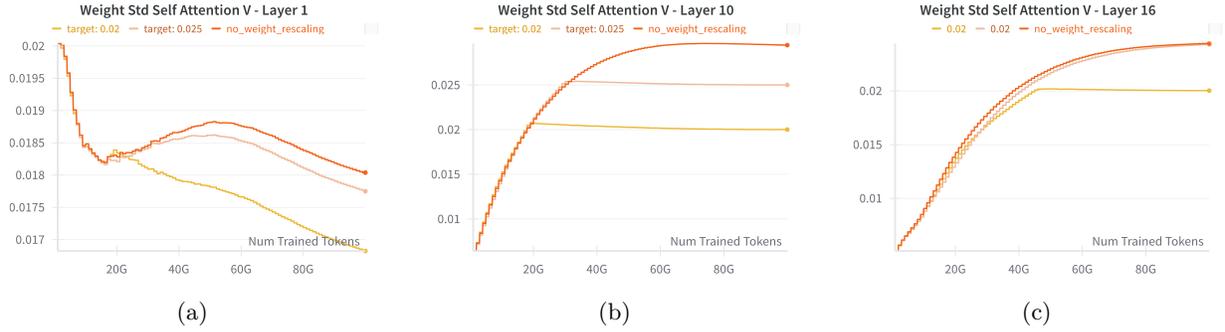

Figure 40: (a) Layer-1 (first); (b) Layer-10; (c) Layer-16 (last); standard deviation evolution (with LIR + TVR of init std 0.02) of Attention V projection matrices for different init standard deviation values.

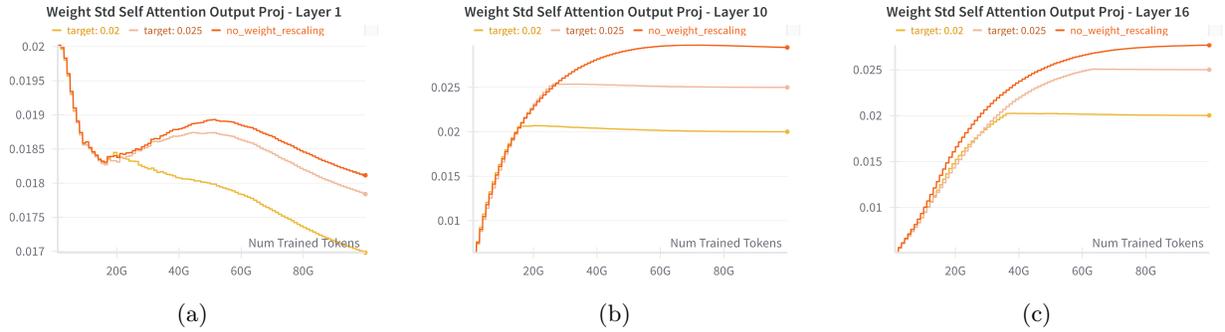

Figure 41: (a) Layer-1 (first); (b) Layer-10; (c) Layer-16 (last); standard deviation evolution (with LIR + TVR of init std 0.02) of Attention Output projection matrices for different init standard deviation values.

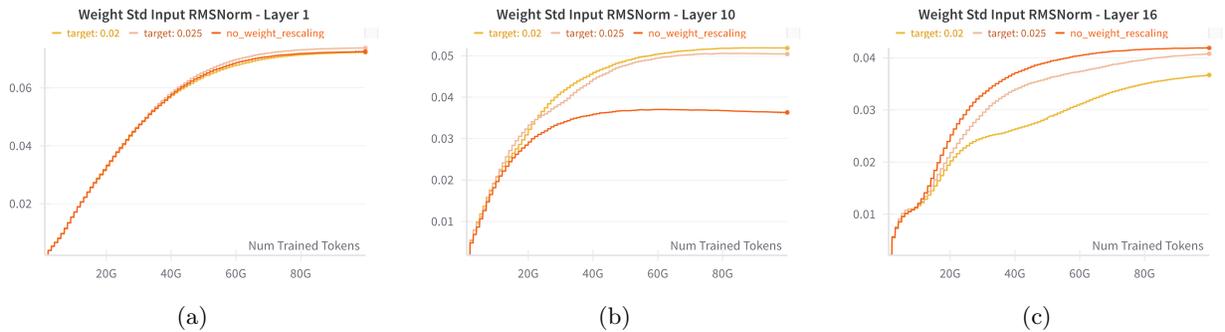

Figure 42: (a) Layer-1 (first); (b) Layer-10; (c) Layer-16 (last); standard deviation evolution (with LIR + TVR of init std 0.02) of Input RMSNorm weight vectors for different init standard deviation values.

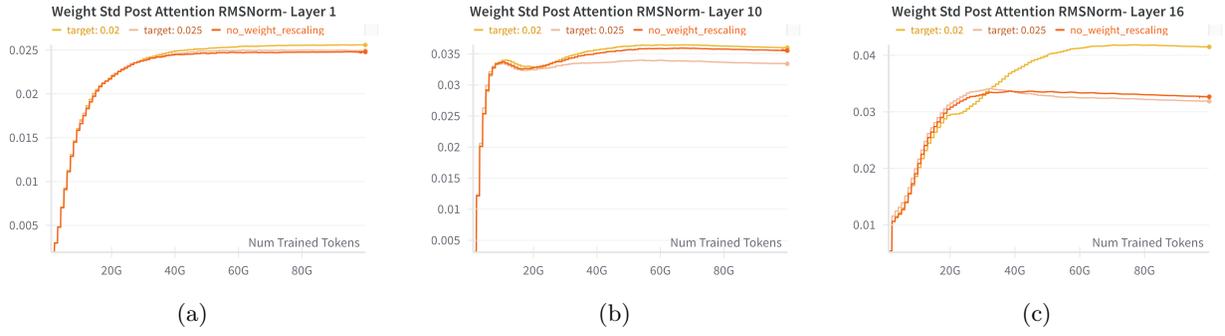

Figure 43: (a) Layer-1 (first); (b) Layer-10; (c) Layer-16 (last); standard deviation evolution (with LIR + TVR of init std 0.02) of Post Attention RMSNorm weight vectors for different init standard deviation values.

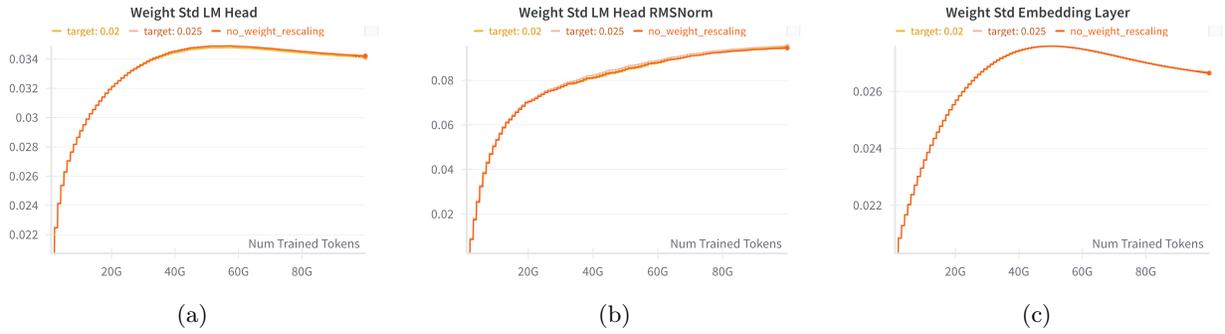

Figure 44: (a) LM-Head; (b) LM-Head RMSNorm; (c) Embedding Layer; standard deviation evolution (with LIR + TVR of init std 0.02) for different init standard deviation values.

RMSNorm, embedding layer, and LMHead layers. Colors in each plot represent different target standard deviation values $\sigma_{\text{target}}$ of the **TVR** applied on 2D modules in the decoder layers as presented in Figure 45 - Figure 54.

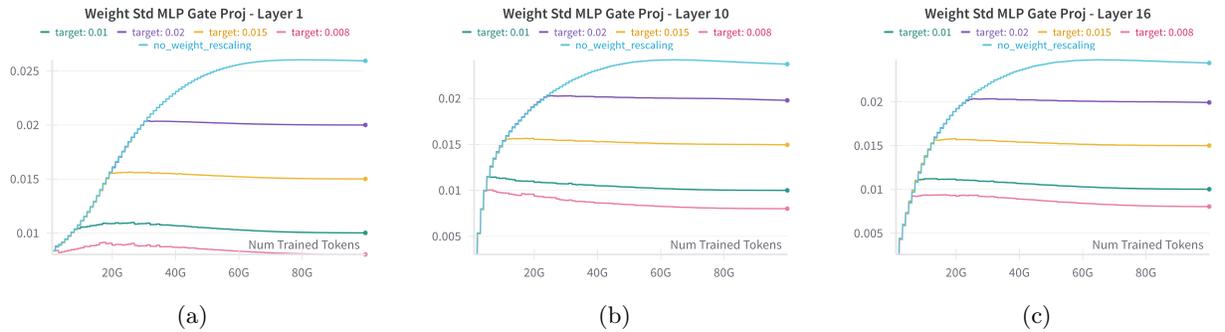

Figure 45: (a) Layer-1 (first); (b) Layer-10; (c) Layer-16 (last); standard deviation evolution (with LIR + TVR of init std 0.008) of MLP Gate Projection matrices for different init standard deviation values.

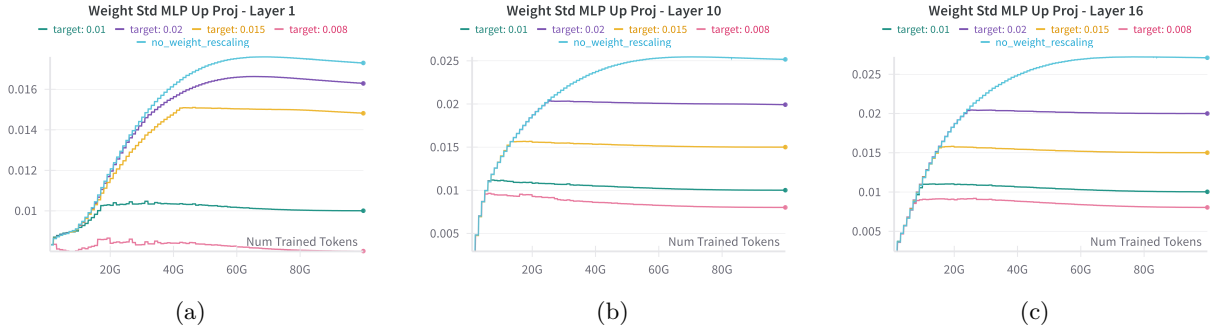

Figure 46: (a) Layer-1 (first); (b) Layer-10; (c) Layer-16 (last); standard deviation evolution (with LIR + TVR of init std 0.008) of MLP Up Projection matrices for different init standard deviation values.

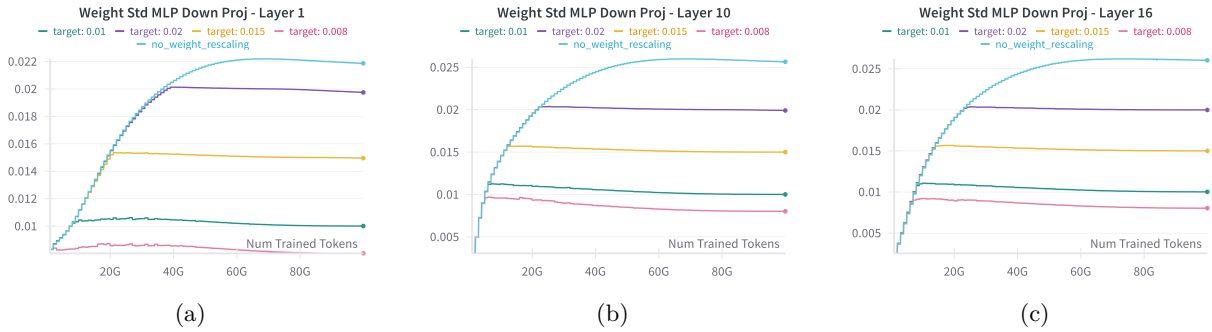

Figure 47: (a) Layer-1 (first); (b) Layer-10; (c) Layer-16 (last); standard deviation evolution (with LIR + TVR of init std 0.008) of MLP Down Projection matrices for different init standard deviation values.

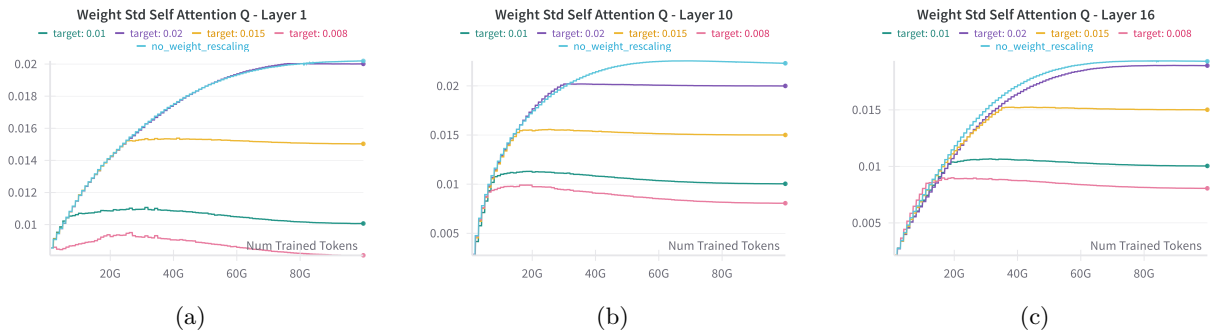

Figure 48: (a) Layer-1 (first); (b) Layer-10; (c) Layer-16 (last); standard deviation evolution (with LIR + TVR of init std 0.008) of Attention Q projection matrices for different init standard deviation values.

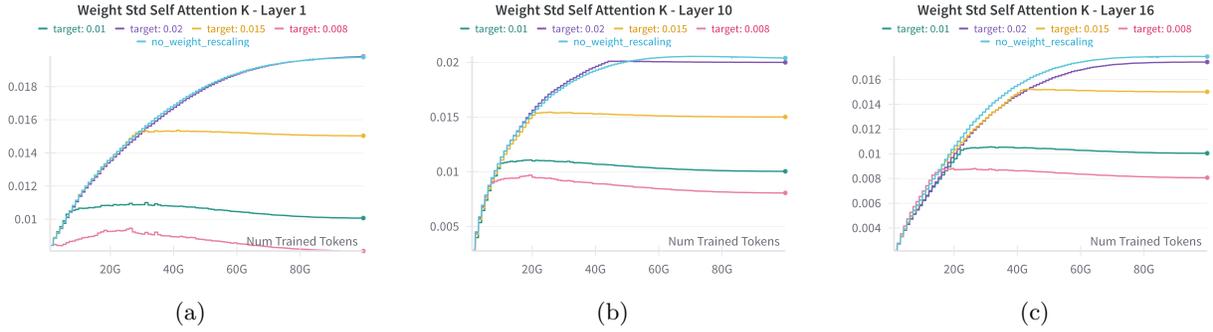

Figure 49: (a) Layer-1 (first); (b) Layer-10; (c) Layer-16 (last); standard deviation evolution (with LIR + TVR of init std 0.008) of Attention K projection matrices for different init standard deviation values.

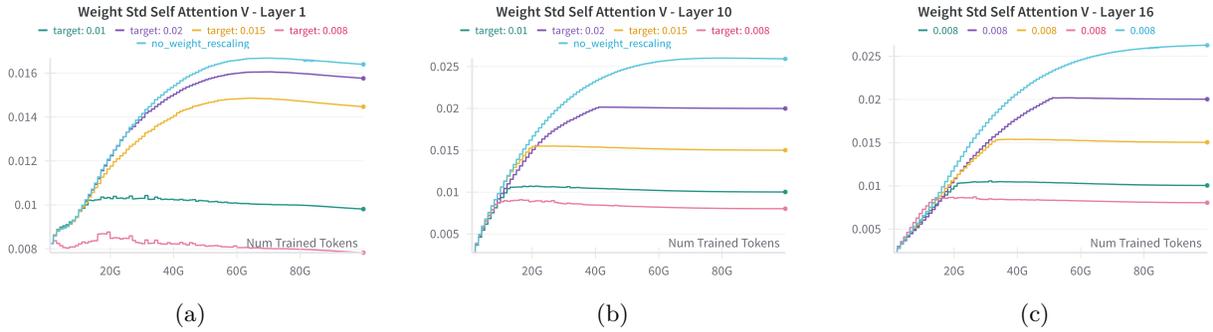

Figure 50: (a) Layer-1 (first); (b) Layer-10; (c) Layer-16 (last); standard deviation evolution (with LIR + TVR of init std 0.008) of Attention V projection matrices for different init standard deviation values.

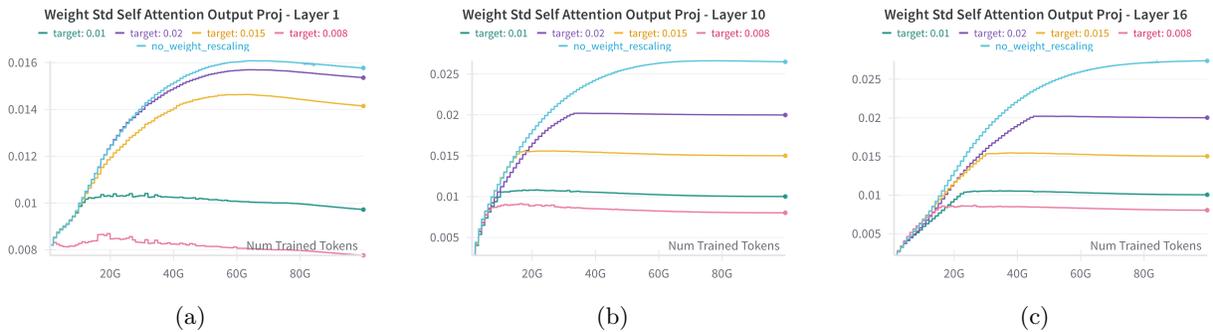

Figure 51: (a) Layer-1 (first); (b) Layer-10; (c) Layer-16 (last); standard deviation evolution (with LIR + TVR of init std 0.008) of Attention Output projection matrices for different init standard deviation values.

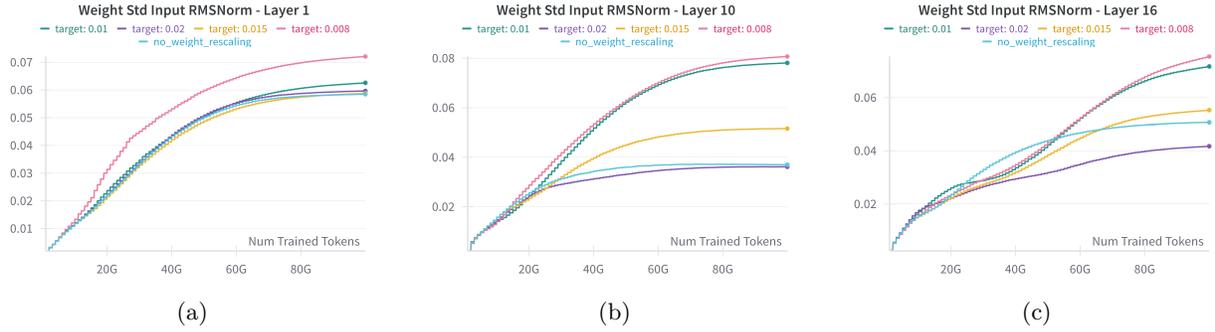

Figure 52: (a) Layer-1 (first); (b) Layer-10; (c) Layer-16 (last); standard deviation evolution (with LIR + TVR of init std 0.008) of Input RMSNorm weight vectors for different init standard deviation values.

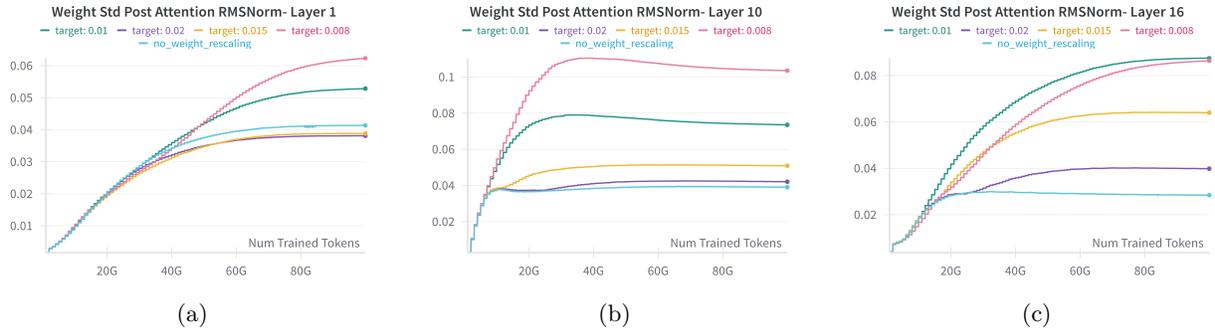

Figure 53: (a) Layer-1 (first); (b) Layer-10; (c) Layer-16 (last); standard deviation evolution (with LIR + TVR of init std 0.008) of Post Attention RMSNorm weight vectors for different init standard deviation values.

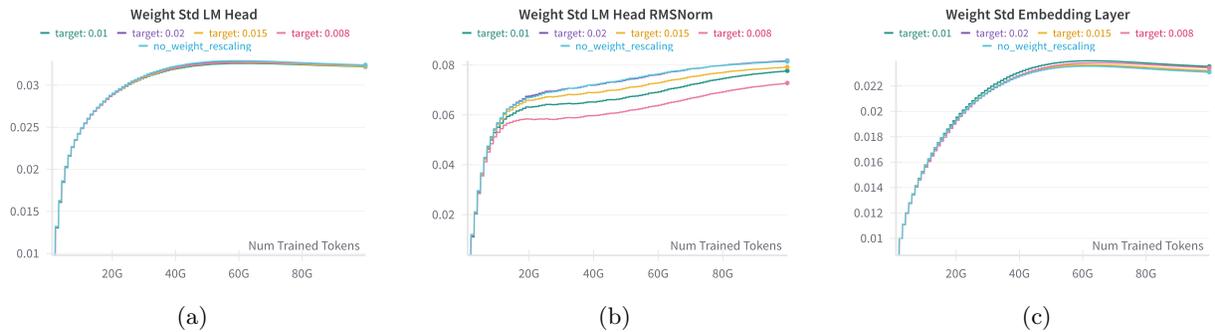

Figure 54: (a) LM-Head; (b) LM-Head RMSNorm; (c) Embedding Layer; standard deviation evolution (with LIR + TVR of init std 0.008) for different init standard deviation values.

### 6.3.5 With Layer Index Rescaling and With Weight Rescaling (Initial Standard Deviation: 0.006

Plots of standard deviation evolution with applied LIR and TVR of all 2D modules in the decoder layers. The plots includes all the 2D modules in the decoder layers as well as the input RMSNorm, post-attention RMSNorm, embedding layer, and LMHead layers. Colors in each plot represent different target standard deviation values $\sigma_{\text{target}}$ of the TVR applied on 2D modules in the decoder layers as presented in Figure 55 - Figure 64.

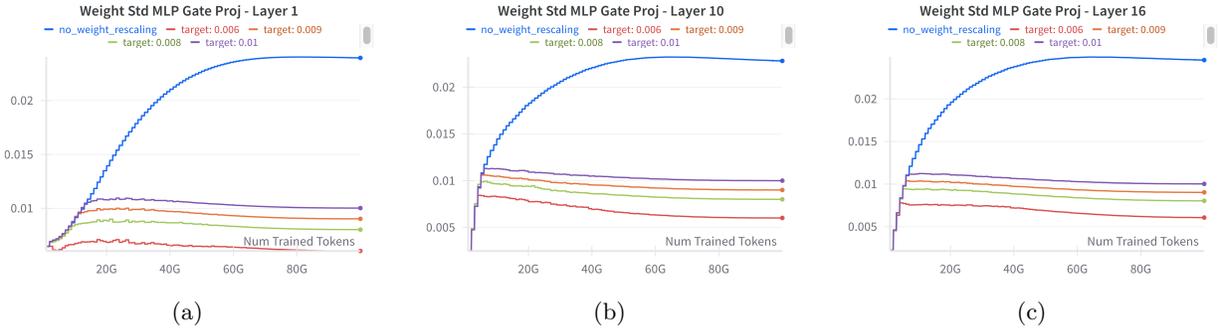

Figure 55: (a) Layer-1 (first); (b) Layer-10; (c) Layer-16 (last); standard deviation evolution (with LIR + TVR of init std 0.006) of MLP Gate Projection matrices for different init standard deviation values.

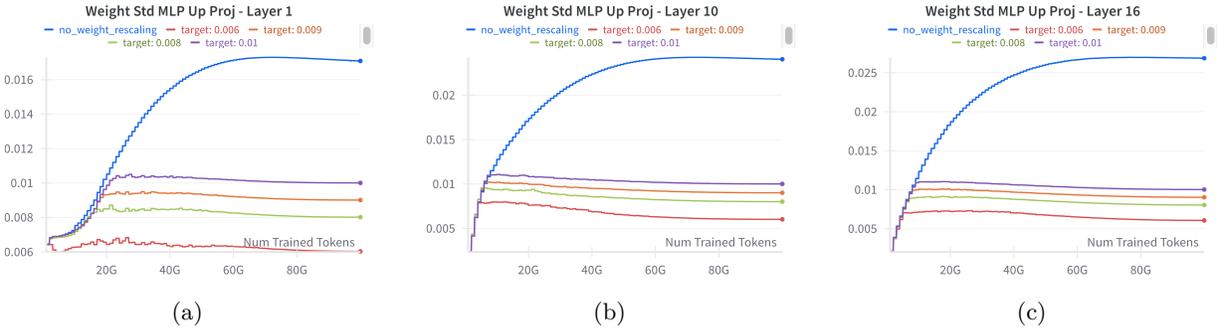

Figure 56: (a) Layer-1 (first); (b) Layer-10; (c) Layer-16 (last); standard deviation evolution (with LIR + TVR of init std 0.006) of MLP Up Projection matrices for different init standard deviation values.

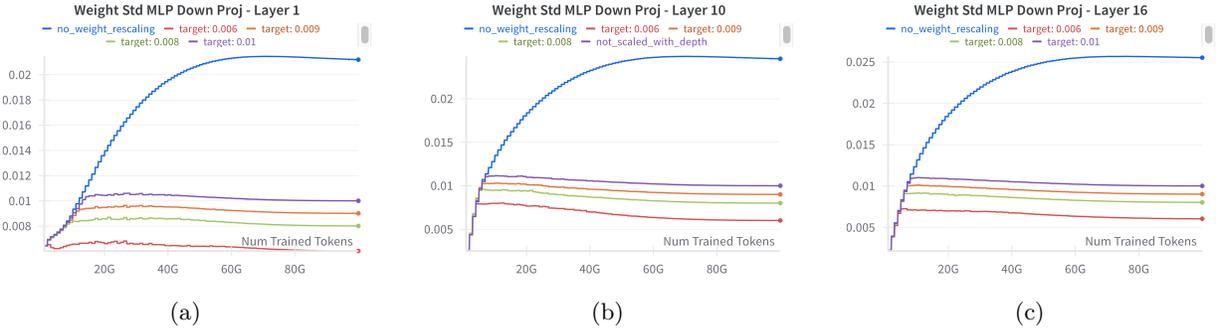

Figure 57: (a) Layer-1 (first); (b) Layer-10; (c) Layer-16 (last); standard deviation evolution (with LIR + TVR of init std 0.006) of MLP Down Projection matrices for different init standard deviation values.

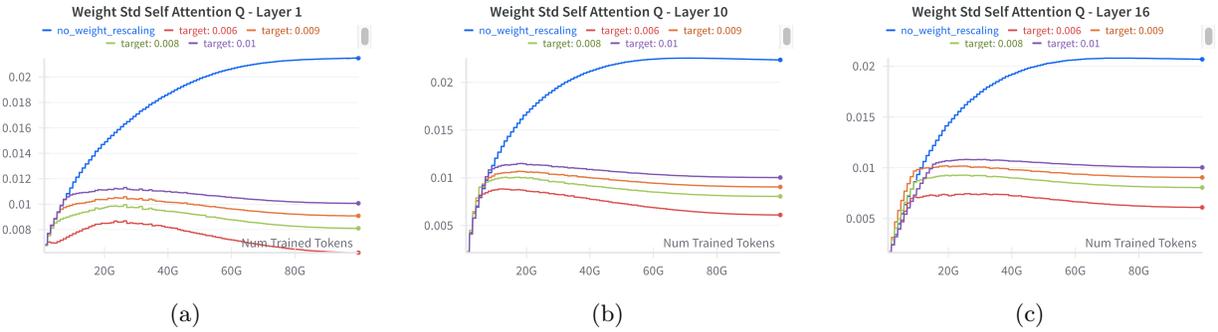

Figure 58: (a) Layer-1 (first); (b) Layer-10; (c) Layer-16 (last); standard deviation evolution (with LIR + TVR of init std 0.006) of Attention Q projection matrices for different init standard deviation values.

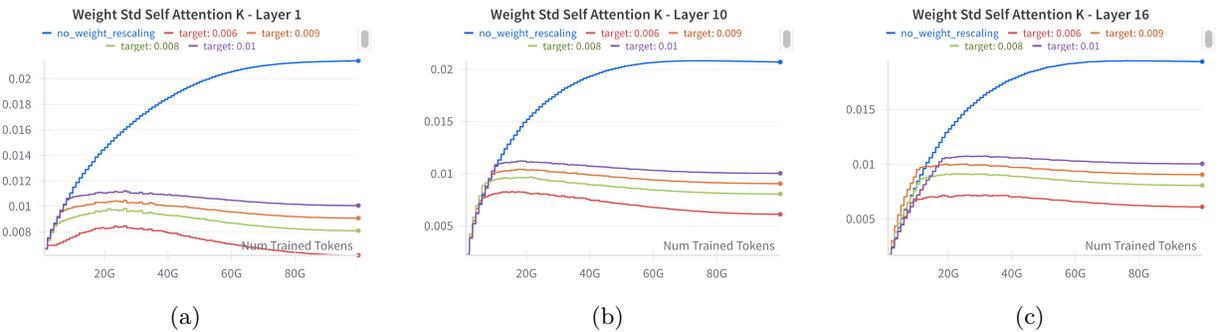

Figure 59: (a) Layer-1 (first); (b) Layer-10; (c) Layer-16 (last); standard deviation evolution (with LIR + TVR of init std 0.006) of Attention K projection matrices for different init standard deviation values.

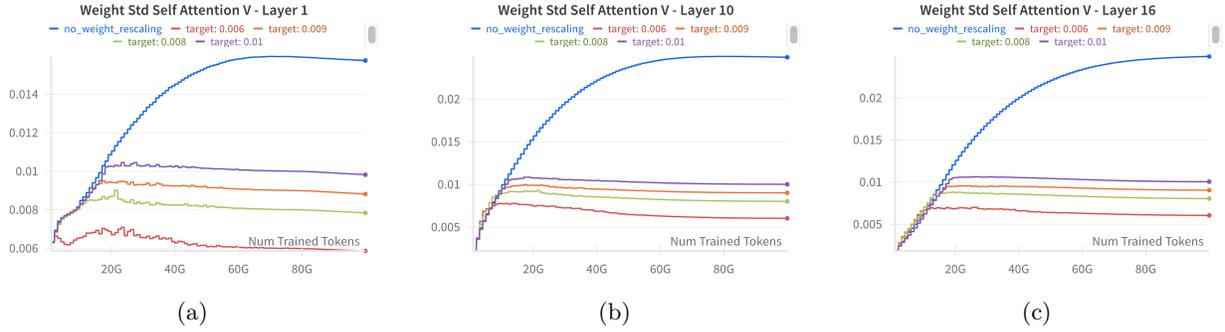

Figure 60: (a) Layer-1 (first); (b) Layer-10; (c) Layer-16 (last); standard deviation evolution (with LIR + TVR of init std 0.006) of Attention V projection matrices for different init standard deviation values.

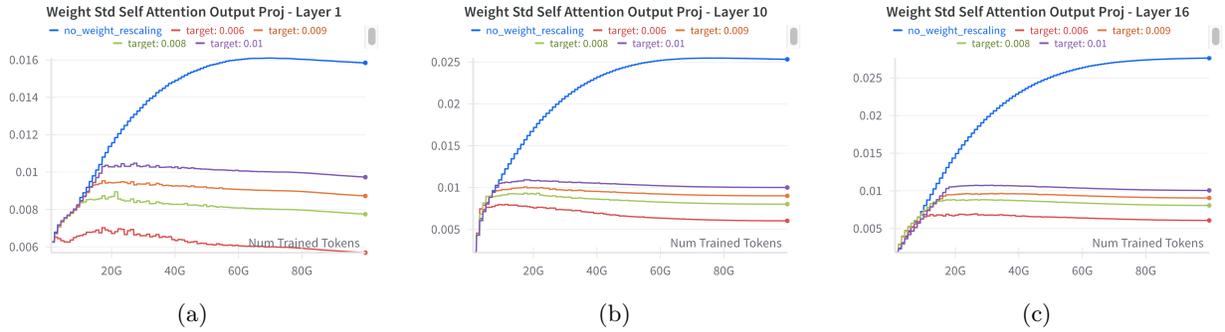

Figure 61: (a) Layer-1 (first); (b) Layer-10; (c) Layer-16 (last); standard deviation evolution (with LIR + TVR of init std 0.006) of Attention Output projection matrices for different init standard deviation values.

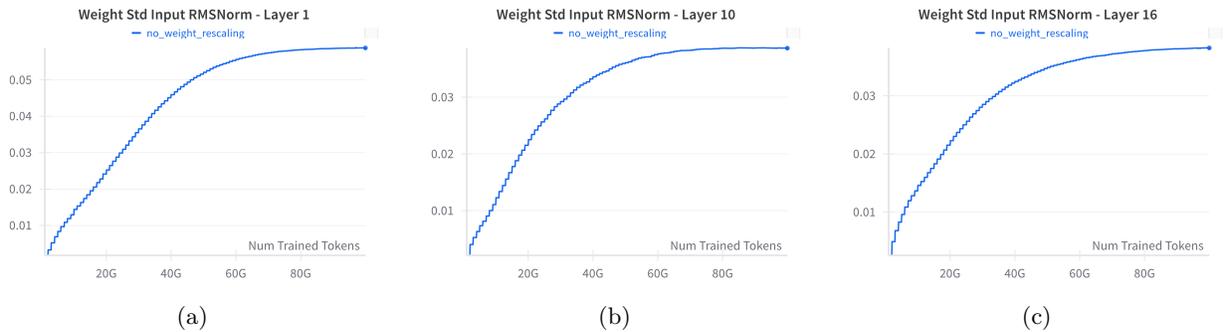

Figure 62: (a) Layer-1 (first); (b) Layer-10; (c) Layer-16 (last); standard deviation evolution (with LIR + TVR of init std 0.006) of Input RMSNorm weight vectors for different init standard deviation values.

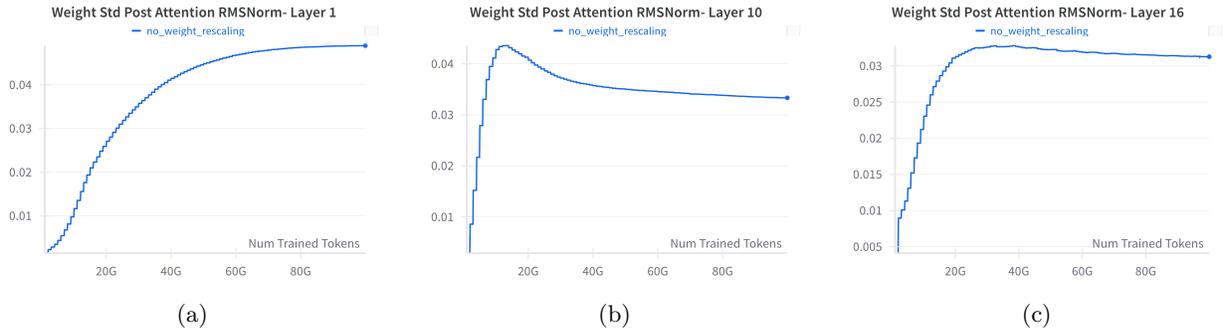

Figure 63: (a) Layer-1 (first); (b) Layer-10; (c) Layer-16 (last); standard deviation evolution (with LIR + TVR of init std 0.006) of Post Attention RMSNorm weight vectors for different init standard deviation values.

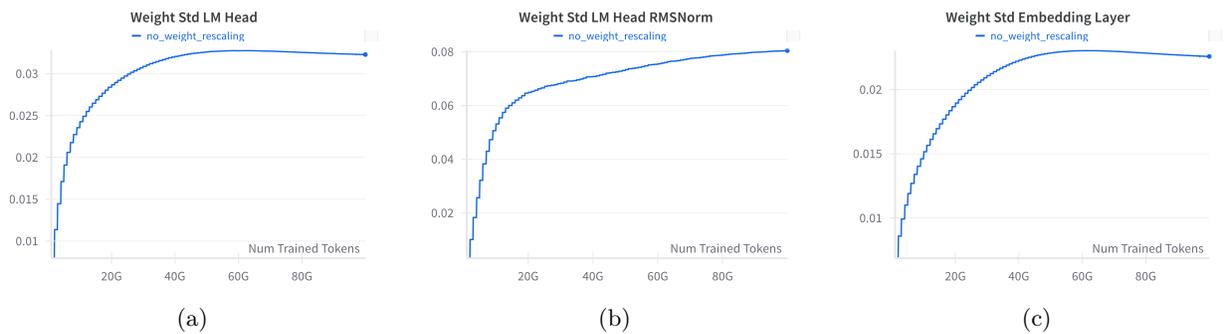

Figure 64: (a) LM-Head; (b) LM-Head RMSNorm; (c) Embedding Layer; standard deviation evolution (with LIR + TVR of init std 0.006) for different init standard deviation values.